\documentclass{article}
\usepackage{log_2026}						% for camera-ready version

\usepackage{booktabs}						% professional-quality tables
\usepackage{multirow}						% tabular cells spanning multiple rows
\usepackage{amsfonts}						% blackboard math symbols
\usepackage{graphicx}						% figures
\usepackage{duckuments}						% sample images

\usepackage{subcaption}
\usepackage{amsmath}
\usepackage{amssymb}
\usepackage{amsthm}
\usepackage{adjustbox}
\usepackage{mathtools}
\usepackage{algorithm}
\usepackage{algpseudocode}

\usepackage[numbers,compress,sort]{natbib}	% for numerical citations
\usepackage{amsmath,amsfonts,bm}

\def\eqref#1{equation~\ref{#1}}
\def\1{\bm{1}}

\DeclareMathAlphabet{\mathsfit}{\encodingdefault}{\sfdefault}{m}{sl}
\SetMathAlphabet{\mathsfit}{bold}{\encodingdefault}{\sfdefault}{bx}{n}

\DeclareMathOperator*{\argmin}{arg\,min}

\def\eqref#1{(\ref{#1})}

\theoremstyle{plain}
\newtheorem{proposition}{Proposition}
\newtheorem{corollary}[proposition]{Corollary}

\theoremstyle{definition}
\newtheorem{definition}{Definition}
\newtheorem{remark}{Remark}
\newtheorem{assumptionT}{Assumption}
\newenvironment{assumption}[1]
  {\begin{assumptionT}(#1).}
  {\end{assumptionT}}

\newcommand{\Gc}{G_{c}}
\newcommand{\Gs}{G_{s}}
\newcommand{\Ghc}{\widehat{G}_{c}}
\newcommand{\Etr}{\mathcal{E}_{\mathrm{tr}}}
\newcommand{\Eall}{\mathcal{E}_{\mathrm{all}}}
\newcommand{\one}{\mathbf{1}}
\newcommand{\Pmat}{\mathbf{P}}
\newcommand{\indep}{\perp\!\!\!\perp}
\title{PISA: Prioritized Invariant Subgraph Aggregation for Out-of-Distribution Generalization on Graphs}

\author[Ghasemi et al.]{%
Ali Ghasemi\textsuperscript{1}
\quad
Farooq Ahmad Wani\textsuperscript{1}
\quad
Maria Sofia Bucarelli\textsuperscript{2}
\quad
Fabrizio Silvestri\textsuperscript{1}
\\[0.5em]
\begin{tabular}{@{}c@{\hspace{2.5em}}c@{}}
\textsuperscript{1}Sapienza University of Rome
&
\textsuperscript{2}Université Côte d’Azur, CNRS, Inria, I3S
\\
\email{\{ghasemi,wani,fsilvestri\}@diag.uniroma1.it}
&
\email{maria-sofia.bucarelli@inria.fr}
\end{tabular}
}

\begin{document}

\maketitle

\begin{abstract}
Invariant learning on graphs aims to build predictors that rely on causal substructures rather than on environment-specific shortcuts. Current methods extract either a \emph{single} invariant subgraph (CIGA) or, more recently, a \emph{set} of diverse invariant subgraphs (SuGAr). We observe that the second half of the multi-subgraph pipeline, "how the extracted subgraphs are \emph{combined}" has gone essentially unexamined; existing methods average branch predictions uniformly or select one greedily. We show that this is precisely where accuracy is lost. A minimum-variance combination argument establishes that uniform averaging is optimal only when every branch carries the same total error covariance with the ensemble---a knife edge balance that second order exchangeability guarantees and that the diversity regularizer used to obtain multiple subgraphs is explicitly designed to break. We then show that invariance is closed under post composition; because the branches are frozen and read the graph only through their invariant subgraphs, \emph{any} measurable aggregator of their outputs remains invariant, and the aggregate's environment gap is bounded by the branches' residual invariance violation \emph{uniformly over aggregators}. Learning the combination rule by plain empirical risk minimization is therefore safe, and aggregator capacity trades off against estimation error rather than robustness. \textbf{PISA} instantiates this with a lightweight instance-conditional MLP over stacked branch logits, trained on frozen branches at negligible cost. Across 15 synthetic and real benchmarks, PISA improves the average over the strongest prior method by $2.96$ points on synthetic shifts and $1.74$ points on realistic shifts, with per-dataset gains up to $5.90$ points, and nearly doubles SuGAr's improvement over ERM on the six DrugOOD tasks (from $+2.76$ to $+5.07$ ROC-AUC).
\end{abstract}

\section{Introduction}

Graph neural networks (GNNs) have become the default tool for learning on relational data, from molecular property prediction to social network analysis \citep{gcn,graphsage,gat,jk,powerful}. Nearly all of them are trained under the assumption that training and test graphs are drawn from the same distribution---an assumption that real deployments routinely violate \citep{ogb,wilds,tdc,drugood}. Shifts introduced by data collection, preprocessing, or the underlying generative process degrade accuracy sharply \citep{terraincognita,shortcuts}, and on graphs they are unusually varied: they act on node attributes \emph{and} on topology, changing size, density, or homophily \citep{local,sizeinvariant,li2022recent}.

\begin{figure}[t]
    \centering
    \begin{minipage}{0.49\linewidth}
        \begin{minipage}[c][4cm][c]{\linewidth}
            \centering
            \includegraphics[width=\linewidth]{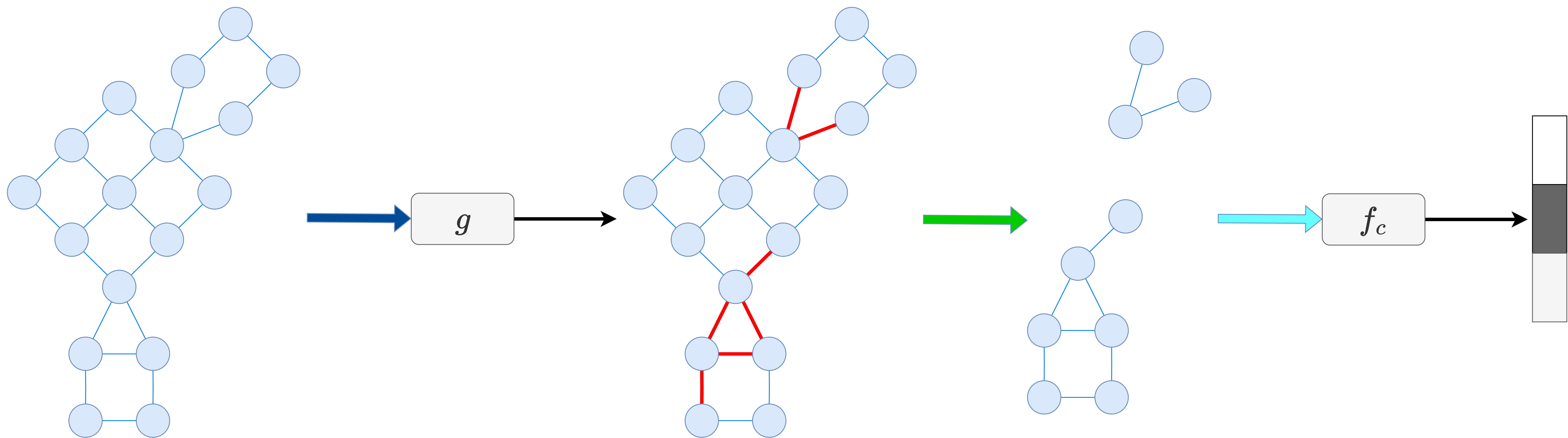}
        \end{minipage}
        \caption{CIGA \citep{ciga}: a featurizer extracts a \emph{single} invariant subgraph, which the classifier uses for prediction.}
        \label{fig:ciga}
    \end{minipage}
    \hfill
    \begin{minipage}{0.49\linewidth}
        \begin{minipage}[c][4cm][c]{\linewidth}
            \centering
            \includegraphics[width=\linewidth]{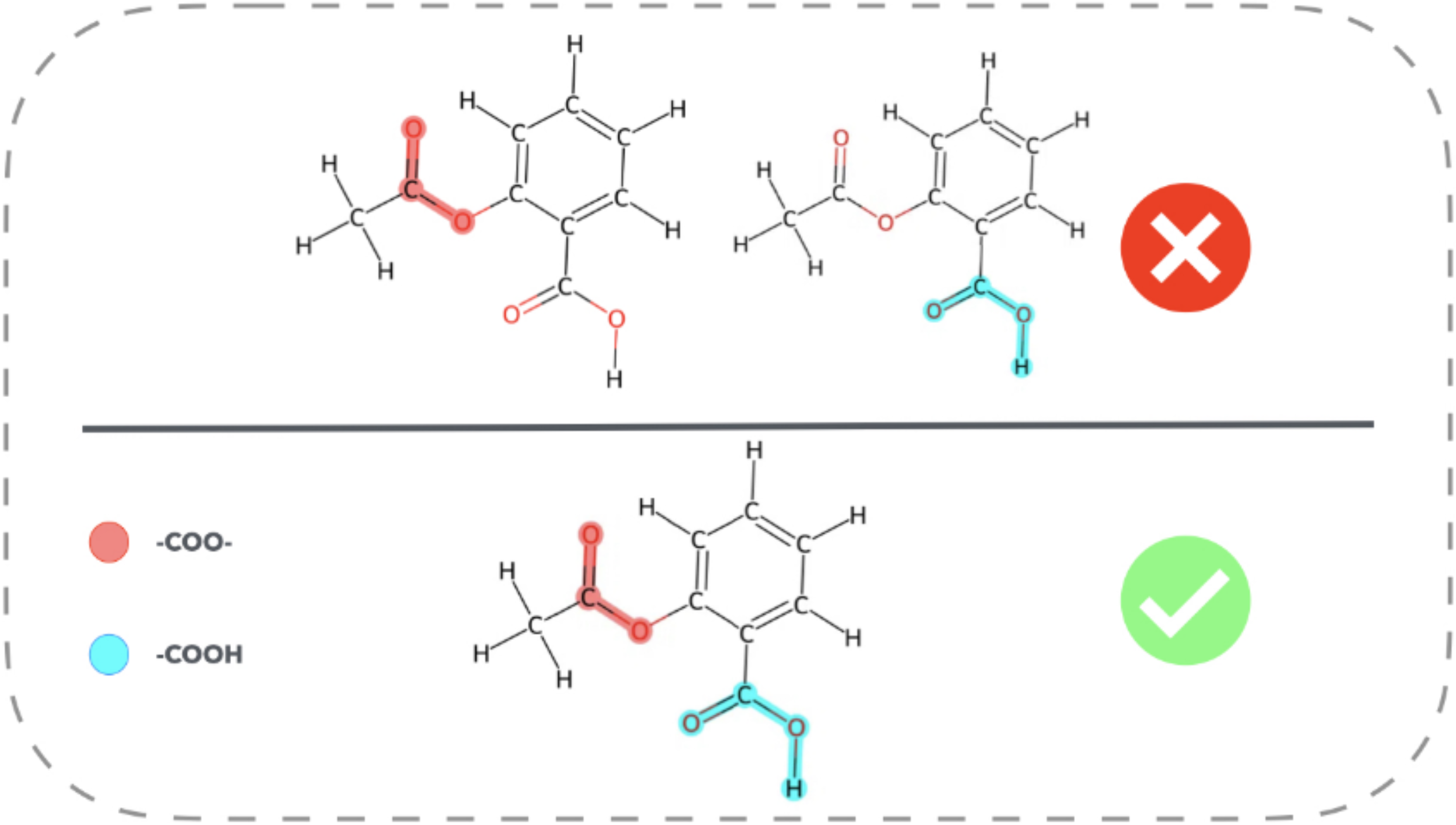}
        \end{minipage}
        \caption{Aspirin's activity depends on \emph{several} functional groups. SuGAr \citep{sugar} extracts multiple subgraphs.}
        \label{fig:multi_causal}
    \end{minipage}
    \vspace{-1.2em}
\end{figure}

The invariance principle offers a principled response. If some part of the input is causally responsible for the label and stable across environments, a predictor restricted to that part generalizes out-of-distribution (OOD) \citep{irm,ibirm,nagarajan2020understanding}. Transferring this idea to graphs requires deciding \emph{which part}. CIGA \citep{ciga} answers with a subgraph: it models graph generation with a structural causal model (SCM) \citep{causality} in which an invariant subgraph $\Gc$ carries all label information, and decomposes a GNN into a featurizer $g$ that proposes $\Ghc$ and a classifier $f_c$ that predicts from it (Fig.~\ref{fig:ciga}). Because environment labels are rarely available for graphs, CIGA supervises $g$ with an information-theoretic surrogate rather than with explicit domain annotations.

Committing to one subgraph, however, misstates many real problems. The activity of a molecule is typically determined by a \emph{combination} of functional groups (Fig.~\ref{fig:multi_causal}); forcing a single mask to explain the label invites it to absorb spurious structure. SuGAr \citep{sugar} addresses this by training $n$ invariant GNNs in parallel and adding a diversity regularizer that pushes their masks apart, so that different branches lock onto different causal substructures. Its final prediction, however, is obtained by \emph{uniform averaging} of the branch outputs, or by greedily keeping the single best branch.

\paragraph{Our starting observation.} The extraction step has received sustained attention; the combination step has not. This is unfortunate, because the two interact in a way that undermines the standard choice. Uniform averaging is the minimum-variance combination precisely when every branch carries the same total error covariance with the ensemble---a balance guaranteed by second-order exchangeability (equal variances and equal pairwise covariances). The diversity regularizer that makes multi-subgraph extraction work is an explicit mechanism for \emph{destroying} that exchangeability: a branch specialized to a motif that happens to be absent from a given graph is unreliable \emph{on that graph}, while its neighbours are not. The optimal combination weights are therefore both non-uniform and \emph{instance-dependent}, and no fixed rule---uniform, greedy, or otherwise---can attain them (Sec.~\ref{sec:bottleneck}).

\paragraph{Why a learned aggregator is safe.} Learning the combination rule invites the objection that a module fitted by empirical risk minimization (ERM) on biased environments will relearn the very shortcuts the branches avoid. It does not apply here. The branches are frozen and see the graph only through their invariant subgraphs, so their outputs are measurable functions of the invariant latent content; invariance is closed under post-composition, making \emph{every} measurable aggregator of those outputs invariant (Prop.~\ref{prop:closure}). Under approximate invariance, the aggregate's environment gap is bounded by the branches' residual violation \emph{uniformly over the aggregator class} (Prop.~\ref{prop:robust}). Capacity therefore trades off against estimation error, not robustness---exactly the behaviour our depth ablation shows.

\paragraph{PISA} We study the following problem: given $n$ invariant predictors extracted by an upstream multi-subgraph method, how should their predictions be combined for OOD generalization? Existing approaches use fixed aggregation, such as uniform averaging or greedy selection. PISA addresses this aggregation problem by freezing the extracted branches and learning a lightweight, instance-conditional MLP over their stacked logits. Our main contribution lies in this aggregation stage and its theoretical analysis, building on existing techniques for invariant-subgraph extraction. Because the aggregator operates on cached branch logits, it requires no additional GNN forward passes during Phase II and adds parameters independent of graph size and backbone width.

\paragraph{Contributions.}
\begin{itemize}\itemsep2pt
    \item We identify \emph{aggregation}, not extraction, as an unexamined bottleneck in multi-subgraph invariant learning, and prove that the diversity mechanism these methods rely on is exactly what makes their uniform-averaging step suboptimal (Prop.~\ref{prop:minvar}, Cor.~\ref{cor:diversity}).
    \item We prove that invariance is preserved under arbitrary measurable aggregation of frozen invariant branches, and give a bound on the environment gap that is uniform over aggregators (Prop.~\ref{prop:closure}, Prop.~\ref{prop:robust}). Together these justify training the aggregator by ERM and predict the capacity/robustness decoupling we observe empirically.
    \item We introduce PISA and evaluate it on 15 benchmarks, improving synthetic and realistic OOD averages by $2.96$ and $1.74$ points over the strongest baseline and nearly doubling SuGAr's gain over ERM on DrugOOD.
\end{itemize}

\section{Preliminaries}
\label{sec:prelim}

\paragraph{Setting.} We study graph-level classification under distribution shift. Let
$\mathcal{D}=\{\mathcal{D}^e\}_{e\in\Eall}$ with $\mathcal{D}^e=\{(G^e_i,Y^e_i)\}_{i=1}^{N_e}$ and $(G^e_i,Y^e_i)\sim\mathbb{P}^e$, where each environment $e$ induces a different joint law over graph--label pairs. A classifier $\rho\circ h$ maps $G$ through an encoder $h:\mathcal{G}\to\mathbb{R}^w$ and a head $\rho:\mathbb{R}^w\to\mathcal{Y}$, with $\mathcal{Y}=\{1,\dots,C\}$. Writing $R^e(\rho\circ h)=\mathbb{E}_{(G,Y)\sim\mathbb{P}^e}[\ell(\rho(h(G)),Y)]$, we observe only $\Etr\subseteq\Eall$ at training time and target the worst case,
\begin{equation}
\min_{\rho,h}\;\max_{e\in\Eall}\;R^e(\rho\circ h).
\label{eq:ood}
\end{equation}

\paragraph{Generative model.} We adopt the latent-variable SCM of \citet{ciga}, stated in full in Appendix~\ref{sec:graph_gen_process}. A latent variable $Z=(C,S)$ splits into invariant \emph{content} $C$ and environment-varying \emph{style} $S$; $C$ generates the invariant subgraph $\Gc$, $S$ generates the spurious part $\Gs$, and the observed graph is $G=f^G_{\mathrm{gen}}(\Gc,\Gs)$. The label is produced by the content alone, $Y\coloneqq f_{\mathrm{inv}}(C)$, while the environment $E$ acts on $S$ only---either through $C$ (FIIF) or through $Y$ (PIIF), or both (MIIF). We make the resulting exogeneity explicit, since our analysis uses it directly.

\begin{assumption}{Exogenous content}\label{assump:exog}
    $C\indep E$, and $Y\coloneqq f_{\mathrm{inv}}(C)$ is a measurable function of $C$ alone.
\end{assumption}

Assumption~\ref{assump:exog} is not an extra requirement: it is implied by the SCMs in Assumptions~\ref{assump:scm_fiif}--\ref{assump:scm_miif}, in which $E$ has no directed path into $C$ and acts only on the spurious branch. It formalizes the intuition that an environment changes assay conditions or scaffold statistics, not the chemistry that determines activity.

\paragraph{Invariant predictors.} CIGA aligns the model with this mechanism: factor the classifier as $f=f_c\circ g$ with a featurizer $g:\mathcal{G}\to\mathcal{G}_c$ proposing $\Ghc$, and supervise $g$ without environment labels through two signals---a supervised contrastive term \citep{khosla2020supervised} that pulls together embeddings of $g(G)$ across same-label graphs (approximating maximization of invariant intra-class mutual information), and a constraint keeping the complement $\widehat{G}_s=G-\Ghc$ less predictive than $\Ghc$, which pushes spurious content out of the selection. In practice $g$ emits dense edge affinities followed by top-$s_c$ selection and $f_c$ is a GNN on the induced subgraph. We call a branch \emph{invariant} when its featurizer succeeds:

\begin{definition}[Invariant branch]\label{def:invariant}
A predictor $f^i=f^i_c\circ g^i$ is \emph{invariant} if its output is a measurable function of the invariant subgraph, i.e.\ $g^i(G)=\gamma^i(\Gc)$ almost surely for some measurable $\gamma^i$.
\end{definition}

\paragraph{Multiple invariant subgraphs.} When the label is determined by several substructures, a single $\Ghc$ is a poor model of the mechanism. Following \citet{sugar} we therefore learn a family of invariant predictors $\{f^i\}_{i=1}^n=\{f^i_c\circ g^i\}_{i=1}^n$, each proposing its own candidate $\Ghc^i$, and denote by
\begin{equation}
\Pmat(G)=\big[p^{(1)};\,\dots;\,p^{(n)}\big]\in\mathbb{R}^{n\times C},
\qquad p^{(i)}=f^i_c\big(g^i(G)\big),
\label{eq:stack}
\end{equation}
the stacked branch outputs. The question this paper addresses is what to do with $\Pmat$.

\section{The Aggregation Bottleneck}
\label{sec:bottleneck}

Existing multi-subgraph methods map $\Pmat$ to a prediction with a fixed rule: uniform averaging, $\widehat{Y}=\frac1n\sum_i p^{(i)}$, or greedy selection of the single validation-best branch. We now show that both are suboptimal, and that the reason is intrinsic to how the branches are produced.

Fix a class coordinate and let $\eta\coloneqq\mathbb{E}[Y\mid \Gc]$ be the Bayes score attainable from the invariant subgraph. Branch $i$ outputs $p^{(i)}=p^{(i)}(G;\theta_i)$, where $\theta_i$ collects the randomness of its Phase~I training (initialization, view sampling, batch order); \emph{all expectations in this section are over $\theta=(\theta_1,\dots,\theta_n)$ with the graph held fixed}, as is standard in ensemble analyses. Write $\varepsilon^{(i)}\coloneqq p^{(i)}-\eta$, assume branches are conditionally unbiased, $\mathbb{E}[\varepsilon^{(i)}\mid G]=0$, and collect the conditional second moments in $\Sigma(G)\coloneqq\mathbb{E}[\varepsilon\varepsilon^{\top}\mid G]\in\mathbb{R}^{n\times n}$, assumed positive definite. Consider combinations $\widehat\eta_w=\sum_i w_i\,p^{(i)}$ preserving unbiasedness, $\one^\top w=1$.

\begin{proposition}[Minimum-variance combination]\label{prop:minvar}
The conditional risk of $\widehat\eta_w$ is $\mathbb{E}[(\widehat\eta_w-\eta)^2\mid G]=w^\top\Sigma(G)\,w$, and is minimized over $\{w:\one^\top w=1\}$ by
\begin{equation}
w^\star(G)=\frac{\Sigma(G)^{-1}\one}{\one^\top\Sigma(G)^{-1}\one},
\qquad
\min_{\one^\top w=1} w^\top\Sigma(G) w=\frac{1}{\one^\top\Sigma(G)^{-1}\one}.
\label{eq:wstar}
\end{equation}
The uniform rule $\bar w=\one/n$ therefore incurs the excess risk
\begin{equation}
\Delta(G)\;=\;\frac{\one^\top\Sigma(G)\one}{n^{2}}-\frac{1}{\one^\top\Sigma(G)^{-1}\one}\;\ge\;0,
\label{eq:excess}
\end{equation}
with equality if and only if $\Sigma(G)\one=\lambda\one$ for some $\lambda>0$.
\end{proposition}

The proof (Appendix~\ref{sec:proofs}) is the classical minimum-variance argument of \citet{bates1969combination}, applied conditionally on the input. The equality condition is what matters: $\Sigma(G)\one\propto\one$ says every branch carries the same \emph{total} error covariance with the ensemble, i.e.\ $\one$ is an eigenvector of $\Sigma(G)$. Second-order exchangeability---identical variances and pairwise covariances---is the canonical way this arises, though sufficient rather than necessary. Either way it is a knife-edge condition, and exactly the regime multi-subgraph methods are engineered to leave.

\begin{corollary}[Diversity defeats uniform averaging]\label{cor:diversity}
Suppose two branches $i\ne j$ have unequal conditional error variances, $\Sigma_{ii}(G)\neq\Sigma_{jj}(G)$, but share their covariances with every other branch, $\Sigma_{ik}(G)=\Sigma_{jk}(G)$ for all $k\notin\{i,j\}$. Then $\Sigma(G)\one\not\propto\one$ and hence $\Delta(G)>0$: uniform averaging is strictly suboptimal.

\emph{Heteroscedasticity alone does not suffice}: unequal variances can be offset by compensating covariances that restore $\Sigma(G)\one\propto\one$ (see the counterexample in Appendix~\ref{sec:proofs}). Cor.~\ref{cor:diversity} isolates the uncompensated case, which is the generic one once branches are pushed onto different substructures.
\end{corollary}

Diversity regularization is such a mechanism: it forces branches onto disjoint substructures, so on a graph where branch $i$'s substructure is present and branch $j$'s absent, $\Sigma_{ii}(G)\ll\Sigma_{jj}(G)$. The dependence on $G$ is essential---competence changes from graph to graph, so $w^\star$ is a \emph{function of the input} and no constant weight vector attains it in expectation. Greedy top-1 selection is no better: it is the corner $w=e_{i^\star}$ of the simplex, optimal only if one branch dominates everywhere.

Conditioning the weights on $G$ would be self-defeating, since a module that reads the graph can read its spurious part too. The next observation is that reading the graph is unnecessary.

\begin{proposition}[Branch outputs suffice]\label{prop:suff}
Identify $Y$ with its one-hot encoding in $\mathbb{R}^{C}$ and let $\mathcal{A}_{\mathrm{all}}$ be the square-integrable measurable maps $\mathbb{R}^{n\times C}\to\mathbb{R}^{C}$. Then $A^\star=\argmin_{A\in\mathcal{A}_{\mathrm{all}}}\mathbb{E}\big[\|A(\Pmat)-Y\|^2\big]$ is the conditional mean $A^\star(\Pmat)=\mathbb{E}[Y\mid\Pmat]$, and $A^\star$ dominates every static rule, since $\Pmat\mapsto w^\top\Pmat$ lies in $\mathcal{A}_{\mathrm{all}}$ for each fixed $w$.
\end{proposition}

The content of Prop.~\ref{prop:suff} is that $\Pmat$ already carries the signal needed to approximate $w^\star(G)$: when branch $i$'s substructure is missing from $G$, its logits are typically low-margin and disagree with the rest, and this is visible in $\Pmat$ without inspecting $G$. A learned aggregator can exploit that; a fixed $w$ cannot. This is stacked generalization \citep{wolpert1992stacked,breiman1996stacked} with an input-conditional gate \citep{jacobs1991adaptive}, specialized to invariant subgraph branches.

\section{PISA}
\label{sec:method}

\begin{figure}[t]
    \begin{subfigure}{0.49\linewidth}
        \centering
        \includegraphics[width=\linewidth]{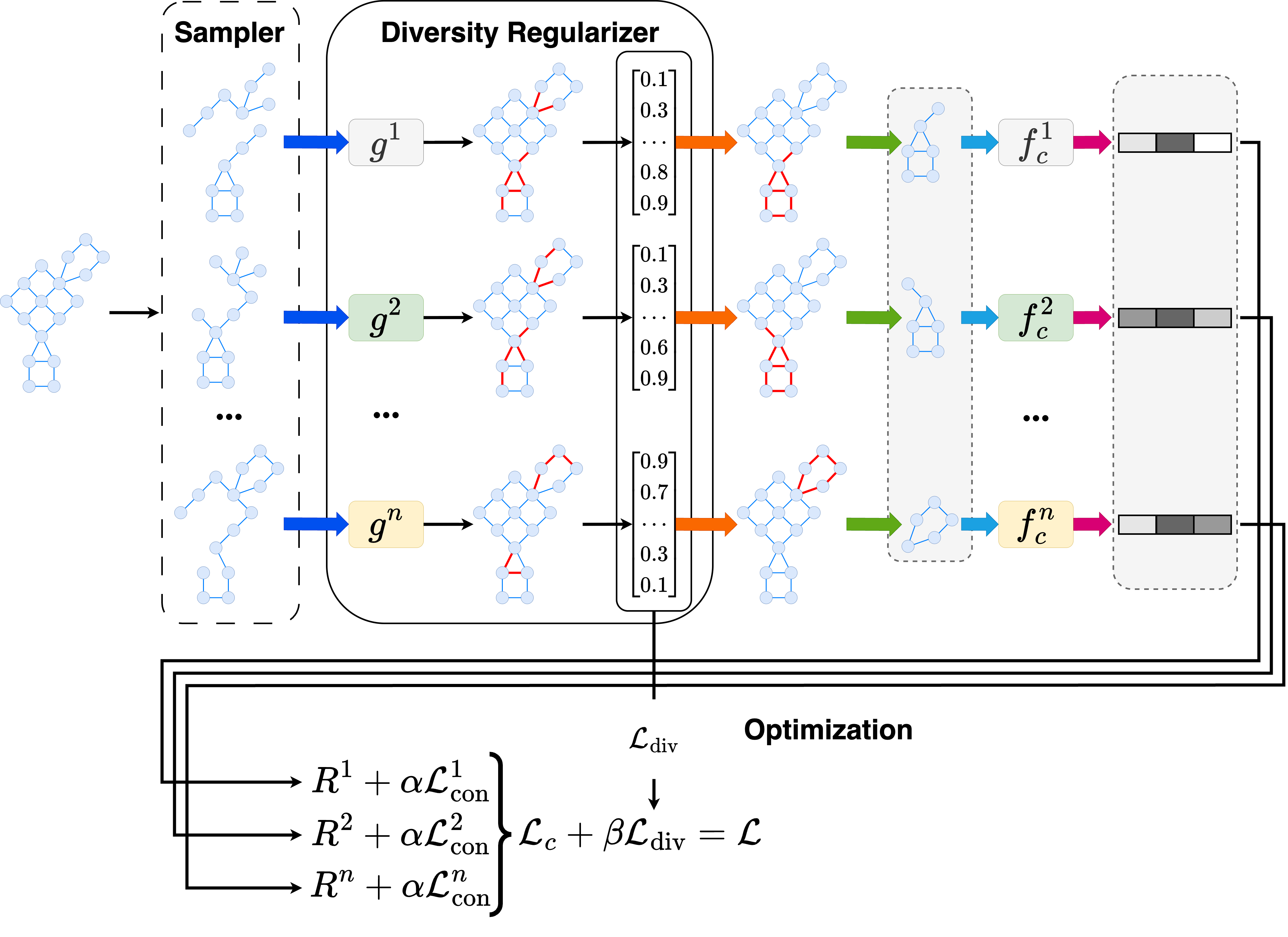}
        \caption{Phase I: diverse invariant subgraph extraction.}
        \label{fig:sugar}
    \end{subfigure}
    \hfill
    \begin{subfigure}{0.49\linewidth}
        \centering
        \includegraphics[width=\linewidth]{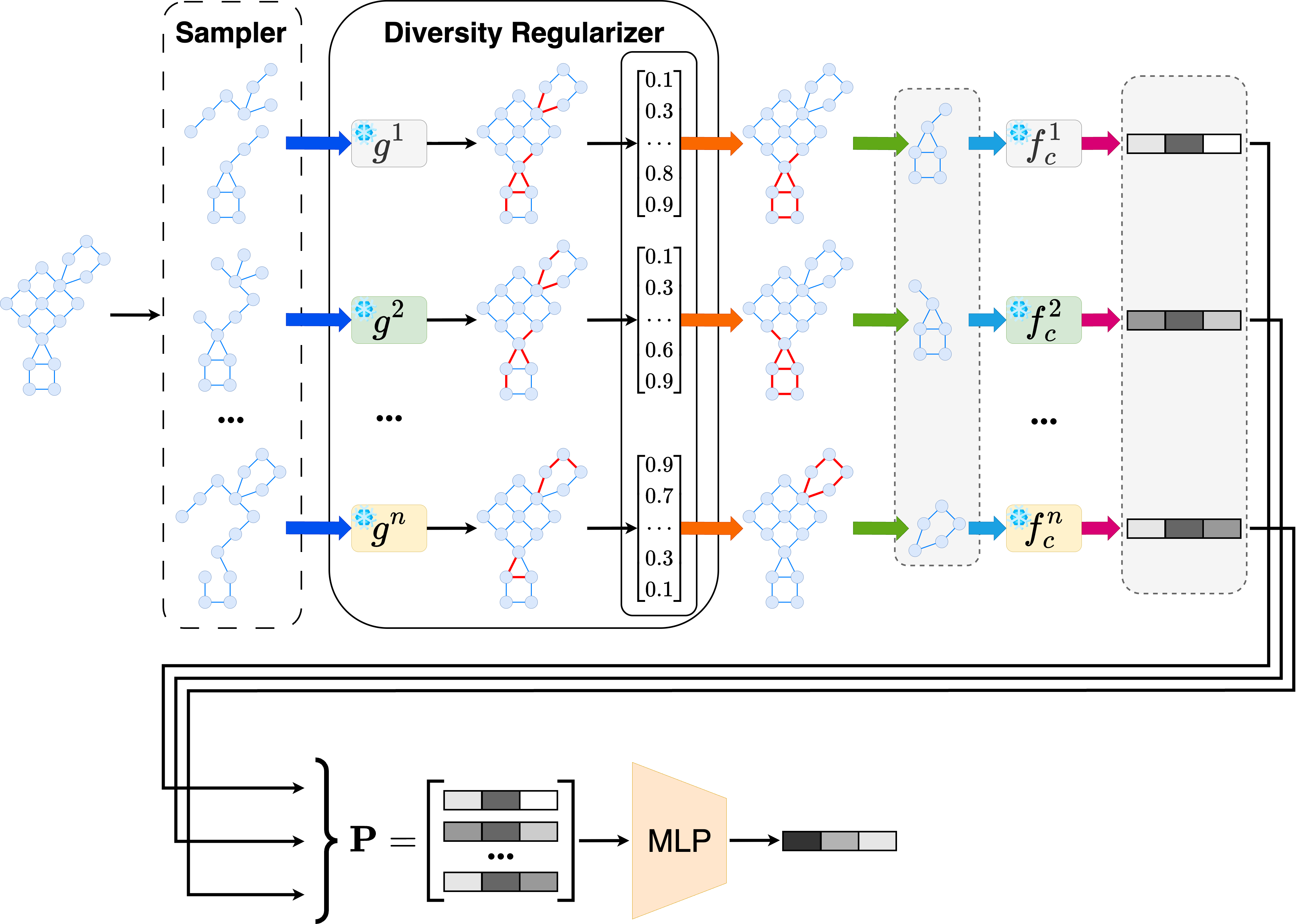}
        \caption{Phase II: prioritized dynamic aggregation.}
        \label{fig:pisa}
    \end{subfigure}
    \caption{\textbf{(a)} A graph sampler stochastically drops edges to create perturbed views. Each view is passed to a featurizer $g^i$ predicting edge-importance weights; a diversity regularizer drives the featurizers towards \emph{distinct} invariant subgraphs $\Ghc^i$, and classifier $f^i_c$ predicts from each. \textbf{(b)} Branch logits are stacked into $\Pmat$ and fed to a lightweight MLP that assigns instance-specific importance to each branch. All $g^i,f^i_c$ are frozen in Phase~II; only the aggregator is trained, with a standard empirical loss.}
    \label{fig:overview}
    \vspace{-0.8em}
\end{figure}

PISA operates on invariant predictors produced by an upstream multi-subgraph extraction stage (Fig.~\ref{fig:overview}). For completeness, Sec.~\ref{subsec:phase1} describes the particular extraction procedure used in our experiments; this stage follows existing invariant-subgraph learning. Sec.~\ref{subsec:phase2} then introduces our contribution: prioritized dynamic aggregation of the frozen branch outputs.

\subsection{Phase I: Upstream Invariant Branch Extraction}\label{subsec:phase1}

\paragraph{Stochastic view sampling.} Each branch receives an independently perturbed view of the input, obtained by dropping edges with probability $q$. Different structural views bias branches towards different substructures and prevent them from converging on a shared shortcut.

\paragraph{Diversity regularization.} Let $m^i=g^i(G)\in[0,1]^{|E|}$ be branch $i$'s soft edge-importance vector. We penalize the pairwise \emph{cosine} overlap of the masks,
\begin{equation}
\delta_{ij}(G)\;=\;\frac{\langle m^i,\,m^j\rangle}{\lVert m^i\rVert_2\,\lVert m^j\rVert_2}\;\in\;[0,1],
\label{eq:div}
\end{equation}
which equals $1$ iff the two branches select proportional masks and $0$ iff their supports are disjoint.

\begin{remark}[Why the normalization matters]\label{rem:degenerate}
Scale-invariance here is not cosmetic. An unnormalized overlap $\langle m^i,m^j\rangle$ is positively homogeneous of degree one in each argument, so a penalty on it can be reduced by uniformly shrinking $\lVert m^i\rVert$ instead of by selecting different edges. Under a fixed top-$s_c$ selection rule the shrunken masks still induce the same subgraphs, so this costs nothing in the forward pass while the diversity signal is quietly lost. Eq.~\eqref{eq:div} closes that escape route.
\end{remark}

\paragraph{Objective.} With $R_{\Ghc^i}$ the supervised loss on branch $i$'s subgraph and $\mathrm{Con}_i$ its supervised-contrastive term,
\begin{equation}
\mathcal{L}_{\mathrm{I}}
=\sum_{i=1}^{n} R_{\Ghc^i}
\;+\;\alpha\sum_{i=1}^{n}\mathrm{Con}_i
\;+\;\beta\sum_{i=1}^{n}\sum_{j\neq i}\delta_{ij}(G),
\label{eq:phase1}
\end{equation}
where, writing $h_{\Ghc}$ for the embedding of subgraph $\Ghc$ and $\phi$ for a similarity function,
\begin{equation}
\mathrm{Con}_i=
\mathbb{E}_{\substack{\{\Ghc^i,\widetilde{G}^i_c\}\sim\mathbb{P}_g(G\mid \mathcal{Y}=Y)\\ \{G_c^{i,k}\}_{k=1}^{M}\sim\mathbb{P}_g(G\mid \mathcal{Y}\neq Y)}}
\log\frac{e^{\phi(h_{\Ghc^i},\,h_{\widetilde{G}^i_c})}}
{e^{\phi(h_{\Ghc^i},\,h_{\widetilde{G}^i_c})}+\sum_{k=1}^{M}e^{\phi(h_{\Ghc^i},\,h_{G_c^{i,k}})}} .
\label{eq:contrastive}
\end{equation}
Here $\widetilde{G}^i_c=g^i(\widetilde{G})$ for a graph $\widetilde{G}$ sharing the label of $G$, and $\alpha,\beta>0$ weight the contrastive and diversity terms.

\subsection{Phase II: Prioritized Dynamic Aggregation}\label{subsec:phase2}

The branches are frozen and their logits stacked as in Eq.~\eqref{eq:stack}. A small MLP $A_\theta$ maps the flattened stack to the final prediction,
\begin{equation}
\widehat{Y}=A_\theta\big(\mathrm{vec}(\Pmat)\big),\qquad A_\theta:\mathbb{R}^{nC}\to\mathbb{R}^{C},
\label{eq:agg}
\end{equation}
trained with the ordinary cross-entropy $\mathcal{L}_{\mathrm{II}}=\mathbb{E}\big[\ell(A_\theta(\mathrm{vec}(\Pmat)),Y)\big]$. Because $A_\theta$ is nonlinear in $\Pmat$, the effective weight it places on branch $i$ varies with the input---this is the ``prioritization'' the method is named for, and by Prop.~\ref{prop:suff} it is the right object to fit.

\paragraph{On permutation symmetry.} A flattened MLP is not permutation-equivariant in the branch index, which would be a defect if branches were exchangeable. They are not: Phase~I ends with $n$ \emph{frozen, individually identifiable} predictors, so branch $i$ means the same thing for every input. Order-sensitivity is usable structure, letting $A_\theta$ learn branch-specific reliability rather than a pooled statistic. A permutation-invariant alternative \citep{zaheer2017deep} is preferable only if $n$ must vary at test time (Sec.~\ref{sec:limitations}).

\subsection{Why Learning the Aggregator Is Safe}
\label{sec:safety}

The obvious worry is that $A_\theta$, fitted by ERM on biased training environments, undoes the invariance that Phase~I paid for. It cannot.

\begin{proposition}[Invariance is closed under aggregation]\label{prop:closure}
Let Assumption~\ref{assump:exog} hold and let every branch be invariant in the sense of Def.~\ref{def:invariant}. Then for \emph{any} measurable $A:\mathbb{R}^{n\times C}\to\mathbb{R}^{C}$, the aggregate $\widehat{Y}=A(\Pmat(G))$ satisfies $(\widehat{Y},Y)\indep E$. Consequently $\mathbb{P}^e(Y\mid\widehat{Y})$ is the same for all $e\in\Eall$, and the risk $R^e(A)$ does not depend on $e$.
\end{proposition}

The argument (Appendix~\ref{sec:proofs}) is short: $\Gc$ is a deterministic function of $C$, hence so is each $p^{(i)}$, hence so is $\widehat{Y}$; $Y$ is a function of $C$ by Assumption~\ref{assump:exog}; and $C\indep E$. The conclusion is unchanged if the branches retain exogenous randomness $U$ (such as the Phase~I edge sampler) as long as $U\indep E$: then $(\widehat{Y},Y)$ is a function of $(C,U)\indep E$. The point is \emph{where} the aggregator sits. It never observes $G$, only the frozen branch outputs, and those have already been stripped of environment-dependent content. No amount of capacity in $A_\theta$ can recover information that is not in its input.

Real branches, of course, are only approximately invariant. The following shows that this is the \emph{only} thing that limits robustness, and that it does so independently of the aggregator.

\begin{proposition}[Uniform environment-gap bound]\label{prop:robust}
Let $\ell$ be bounded by $B$ and let
$\;\delta\coloneqq\sup_{e,e'\in\Eall} d_{\mathrm{TV}}\!\big(\mathbb{P}^e_{(\Pmat,Y)},\,\mathbb{P}^{e'}_{(\Pmat,Y)}\big)$
measure the residual environment-dependence of the branch outputs. Then
\begin{equation}
\sup_{e,e'\in\Eall}\big|R^{e}(A)-R^{e'}(A)\big|\;\le\;2B\delta
\qquad\text{for every measurable }A .
\label{eq:gap}
\end{equation}
If all branches are exactly invariant, $\delta=0$ and Eq.~\eqref{eq:gap} recovers Prop.~\ref{prop:closure}.
\end{proposition}

The bound is \emph{uniform in $A$}: enlarging the aggregator class does not enlarge the worst-case environment gap. Robustness is inherited wholesale from Phase~I. What does depend on $A$ is estimation error. Combining Eq.~\eqref{eq:gap} with any uniform-convergence bound $\mathfrak{C}(\mathcal{A})$ over the aggregator class gives the schematic decomposition
\begin{equation}
R^{e_{\mathrm{te}}}(A)\;\le\;\underbrace{\widehat{R}^{\,\mathrm{tr}}(A)}_{\text{fit}}\;+\;\underbrace{2B\delta}_{\text{shift; independent of }A}\;+\;\underbrace{\mathfrak{C}(\mathcal{A})}_{\text{complexity of }\mathcal{A}} .
\label{eq:decomp}
\end{equation}

\begin{remark}[A prediction about aggregator depth]\label{rem:depth}
Since the middle term of Eq.~\eqref{eq:decomp} is constant in $A$, choosing the aggregator is a pure approximation--estimation trade-off over a \emph{low-dimensional} input: $\mathrm{vec}(\Pmat)\in\mathbb{R}^{nC}$, which is $30$ dimensional for $n=10,C=3$. We should therefore expect a shallow optimum---enough depth to represent instance-conditional gating, after which $\mathfrak{C}(\mathcal{A})$ grows with no compensating gain. Sec.~\ref{sec:ablation_mlp_layers} confirms a $3$--$4$ layer optimum, which is a prediction of the theory rather than a tuned artefact.
\end{remark}

\subsection{Cost}

Algorithm~\ref{alg:pisa} in Appendix~\ref{sec:algo} states the full procedure. Phase~I costs $\mathcal{O}(n\,T\,|E|\,d)$ for $T$ steps and hidden width $d$---identical to SuGAr. Phase~II is where PISA's overhead lives, and it is small: the branch logits are computed \emph{once} and cached as an $N\times n\times C$ tensor, so each subsequent epoch touches no graph structure at all. The aggregator adds $\mathcal{O}(nCd_A+L d_A^2)$ parameters for depth $L$ and width $d_A$, a count that is independent of graph size, of $|E|$, and of the GNN backbone width. PISA's total training cost is thus that of SuGAr plus a term that does not scale with the data's graph structure.

\section{Experiments}

We evaluate PISA on \textbf{15} datasets spanning synthetic and real-world shifts, and organize the results around four questions. \textbf{RQ1:} does PISA improve OOD generalization over the state of the art? \textbf{RQ2:} does the benefit survive when the label genuinely depends on multiple subgraphs? \textbf{RQ3:} how does the number of invariant branches affect performance? \textbf{RQ4:} does aggregator capacity behave as Sec.~\ref{sec:safety} predicts? \textbf{RQ5:} can the branch ensemble be made cheaper by sharing parameters? Datasets, baselines, implementation details, hyperparameters, and the evaluation protocol are provided in Appendix~\ref{sec:app_emp}. All results are mean $\pm$ standard deviation over five independent runs, with model selection based exclusively on validation performance. We report accuracy everywhere except DrugOOD, where we follow \citet{drugood} and report ROC-AUC.

\subsection{Main Results (RQ1)}

\renewcommand{\arraystretch}{1.15}
\begin{table}[t]
    \caption{OOD generalization on synthetic graphs (accuracy \%). Best in \textbf{bold}, second best \underline{underlined}. AVG is the mean over the six columns.}
    \label{tab:results_synthetic}
    \centering
    \resizebox{\linewidth}{!}{
        \begin{tabular}{lccccccc}
            \toprule
            \multirow{2}{*}{Method} &\multicolumn{3}{c}{SPMotif} & \multicolumn{3}{c}{SUMotif} & \multirow{2}{*}{AVG} \\
            \cmidrule(lr){2-4}\cmidrule(lr){5-7}
            & $\text{bias}=0.33$ & $\text{bias}=0.6$ & $\text{bias}=0.9$ & $\text{bias}=0.33$ & $\text{bias}=0.6$ & $\text{bias}=0.9$ & \\
            \midrule
            IRM \citep{irm} & $63.98_{\pm 8.51}$ & $61.58_{\pm 12.85}$ & $47.14_{\pm 12.13}$ & $61.39_{\pm 13.10}$ & $58.48_{\pm 15.47}$ & $48.35_{\pm 14.91}$ & $56.82$ \\
            V-REx \citep{rex}  & $69.18_{\pm 7.34}$ & $58.76_{\pm 11.51}$ & $43.81_{\pm 13.21}$ & $63.24_{\pm 15.63}$ & $65.23_{\pm 14.18}$ & $44.03_{\pm 12.40}$ & $57.38$ \\
            IB-IRM \citep{ibirm} & $62.30_{\pm 11.27}$ & $59.78_{\pm 12.77}$ & $46.19_{\pm 12.10}$ & $71.13_{\pm 11.04}$ & $56.55_{\pm 13.60}$ & $47.27_{\pm 13.31}$ & $57.20$ \\
            EIIL \citep{eiil} & $63.82_{\pm 12.43}$ & $59.42_{\pm 13.16}$ & $42.51_{\pm 11.89}$ & $67.66_{\pm 10.27}$ & $55.64_{\pm 13.11}$ & $41.22_{\pm 8.56}$ & $55.04$ \\
            \midrule
            GREA \citep{grea} & $74.53_{\pm 8.89}$ & $68.26_{\pm 10.53}$ & $48.45_{\pm 13.96}$ & $68.17_{\pm 17.64}$ & $54.39_{\pm 12.37}$ & $51.83_{\pm 15.03}$ & $60.94$ \\
            GSAT \citep{gsat} & $65.51_{\pm 8.54}$ & $56.20_{\pm 6.85}$  & $48.28_{\pm 14.24}$ & $57.81_{\pm 9.29}$ & $65.12_{\pm 5.78}$  & $60.61_{\pm 9.27}$  & $58.92$ \\
            DisC \citep{disc} & $70.01_{\pm 9.84}$ & $54.50_{\pm 13.26}$ & $44.40_{\pm 12.29}$ & $76.20_{\pm 6.94}$ & $62.42_{\pm 18.18}$ & $48.07_{\pm 9.85}$  & $59.27$ \\
            CAL \citep{cal} & $69.51_{\pm 5.38}$ & $64.05_{\pm 5.59}$  & $51.49_{\pm 10.89}$ & $69.60_{\pm 4.20}$ & $53.71_{\pm 8.35}$  & $44.04_{\pm 4.94}$  & $58.73$ \\
            GIL \citep{gil} & $70.79_{\pm 8.48}$ & $71.38_{\pm 11.12}$ & $53.24_{\pm 16.46}$ & $71.43_{\pm 7.96}$ & $64.32_{\pm 13.46}$ & $38.02_{\pm 6.61}$  & $61.53$ \\
            DIR \citep{dir} & $59.58_{\pm 7.86}$ & $66.52_{\pm 7.95}$  & $39.28_{\pm 3.52}$  & $52.59_{\pm 6.35}$ & $45.84_{\pm 6.35}$  & $37.04_{\pm 4.89}$  & $50.14$ \\
            CIGA \citep{ciga} & $63.70_{\pm 8.47}$ & $64.42_{\pm 12.69}$ & $53.20_{\pm 19.19}$ & $64.17_{\pm 12.10}$ & $53.20_{\pm 18.48}$ & $48.28_{\pm 14.24}$ & $57.83$ \\
            \midrule
            CIGA$\times$10 & $67.17_{\pm 7.43}$ & $65.90_{\pm 8.16}$ & $69.53_{\pm 9.62}$ & $77.70_{\pm 8.24}$ & $74.13_{\pm 7.51}$ & $59.30_{\pm 10.43}$ & $68.96$ \\
            SuGAr \citep{sugar} & \underline{$82.82_{\pm 8.90}$} & \underline{$84.57_{\pm 4.38}$}  & \underline{$81.61_{\pm 8.40}$}  & \underline{$78.39_{\pm 9.27}$} & \underline{$79.21_{\pm 5.79}$}  & \underline{$75.87_{\pm 6.33}$}  & \underline{$80.41$} \\
            \textbf{PISA} (ours) & $\boldsymbol{88.72_{\pm 8.10}}$ & $\boldsymbol{87.12_{\pm 5.04}}$ & $\boldsymbol{84.02_{\pm 7.95}}$ & $\boldsymbol{80.77_{\pm 8.60}}$ & $\boldsymbol{81.44_{\pm 5.32}}$ & $\boldsymbol{78.15_{\pm 6.20}}$ & $\boldsymbol{83.37}$ \\
            \bottomrule
        \end{tabular}
    }
    \vspace{-0.4em}
\end{table}

\begin{table}[t]
    \centering
    \caption{OOD generalization under realistic shifts. ROC-AUC for the six DrugOOD tasks (EC-$\ast$, Ki-$\ast$), accuracy elsewhere. Best in \textbf{bold}, second best \underline{underlined}. AVG is the mean over the nine columns.}
    \resizebox{\linewidth}{!}{
        \begin{tabular}{lcccccccccc}
            \toprule
            Method & EC-Assay & EC-Scaffold & EC-Size & SST5 & Twitter & CMNIST & Ki-Assay & Ki-Scaffold & Ki-Size & AVG \\
            \midrule
            ERM \citep{principles} & $75.57_{\pm 1.23}$ & $64.21_{\pm 0.89}$ & $63.30_{\pm 1.19}$ & $\underline{44.21_{\pm 0.91}}$ & \underline{$63.84_{\pm 1.61}$} & $10.26_{\pm 0.62}$ & $73.30_{\pm 1.67}$ & $70.45_{\pm 0.30}$ & $74.00_{\pm 1.55}$ & $59.90$ \\
            IRM \citep{irm} & $77.10_{\pm 2.55}$ & $64.32_{\pm 0.42}$ & $62.33_{\pm 0.86}$ & $42.77_{\pm 1.26}$ & $60.42_{\pm 1.06}$ & $15.15_{\pm 3.66}$ & $75.10_{\pm 3.38}$ & $69.32_{\pm 1.84}$ & $76.25_{\pm 0.73}$ & $60.31$ \\
            V-REx \citep{rex} & $75.57_{\pm 2.17}$ & $64.73_{\pm 0.53}$ & $62.80_{\pm 0.89}$ & $42.48_{\pm 1.67}$ & $60.50_{\pm 2.05}$ & $17.12_{\pm 5.68}$ & $74.16_{\pm 1.46}$ & $71.40_{\pm 2.77}$ & $76.68_{\pm 1.35}$ & $60.60$ \\
            IB-IRM \citep{ibirm} & $64.70_{\pm 2.50}$ & $62.62_{\pm 2.05}$ & $58.28_{\pm 0.99}$ & $43.02_{\pm 1.94}$ & $60.80_{\pm 2.50}$ & $13.06_{\pm 1.97}$ & $71.98_{\pm 3.26}$ & $69.55_{\pm 1.66}$ & $70.71_{\pm 1.95}$ & $57.19$ \\
            EIIL \citep{eiil} & $64.20_{\pm 5.40}$ & $62.88_{\pm 2.75}$ & $59.58_{\pm 0.96}$ & $43.79_{\pm 1.19}$ & $60.15_{\pm 1.44}$ & $11.80_{\pm 0.42}$ & $74.24_{\pm 2.48}$ & $69.63_{\pm 1.46}$ & $76.56_{\pm 1.37}$ & $58.09$ \\
            \midrule
            GREA \citep{grea} & $66.87_{\pm 7.53}$ & $63.14_{\pm 2.19}$ & $59.20_{\pm 1.42}$ & $43.29_{\pm 0.85}$ & $59.92_{\pm 1.48}$ & $13.92_{\pm 3.43}$ & $73.17_{\pm 1.80}$ & $67.82_{\pm 4.67}$ & $73.52_{\pm 2.75}$ & $57.87$ \\
            GSAT \citep{gsat} & $76.07_{\pm 1.95}$ & $63.58_{\pm 1.36}$ & $61.12_{\pm 0.66}$ & $43.24_{\pm 0.61}$ & $60.13_{\pm 1.51}$ & $10.51_{\pm 0.53}$ & $72.26_{\pm 1.76}$ & $70.16_{\pm 0.80}$ & $75.78_{\pm 2.60}$ & $59.21$ \\
            DisC \citep{disc} & $61.94_{\pm 7.76}$ & $54.10_{\pm 5.69}$ & $57.64_{\pm 1.57}$ & $40.67_{\pm 1.19}$ & $57.89_{\pm 2.02}$ & $15.08_{\pm 0.21}$ & $54.12_{\pm 8.53}$ & $55.35_{\pm 10.5}$ & $50.83_{\pm 9.30}$ & $49.74$ \\
            CAL \citep{cal} & $75.10_{\pm 2.71}$ & $64.79_{\pm 1.58}$ & $63.38_{\pm 0.88}$ & $39.60_{\pm 1.80}$ & $55.36_{\pm 2.67}$ & $11.46_{\pm 1.82}$ & $75.10_{\pm 0.73}$ & $60.35_{\pm 11.3}$ & $73.69_{\pm 2.29}$ & $57.65$ \\
            GIL \citep{gil} & $70.56_{\pm 4.46}$ & $61.59_{\pm 3.16}$ & $60.46_{\pm 1.91}$ & $43.30_{\pm 1.24}$ & $61.78_{\pm 1.66}$ & $13.19_{\pm 2.25}$ & \underline{$75.22_{\pm 1.73}$} & $71.08_{\pm 4.83}$ & $72.93_{\pm 1.79}$ & $58.90$ \\
            CIGA \citep{ciga} & \underline{$77.52_{\pm 0.97}$} & $61.76_{\pm 1.13}$ & $63.74_{\pm 1.43}$ & $44.20_{\pm 1.89}$ & $60.94_{\pm 1.04}$ & $10.44_{\pm 0.39}$ & $71.98_{\pm 2.65}$ & $73.98_{\pm 2.37}$ & $77.00_{\pm 2.36}$ & $60.17$ \\
            \midrule
            CIGA$\times$10 & $77.46_{\pm 1.23}$ & $62.25_{\pm 1.14}$ & $63.63_{\pm 1.31}$ & \boldsymbol{$44.91_{\pm 1.42}$} & $62.86_{\pm 1.73}$ & $13.55_{\pm 6.82}$ & $73.92_{\pm 2.41}$ & $74.45_{\pm 2.13}$ & $78.29_{\pm 1.92}$ & $61.26$ \\
            SuGAr \citep{sugar} & $76.25_{\pm 1.43}$ & \underline{$65.27_{\pm 1.17}$} & \underline{$64.53_{\pm 2.72}$} & $43.38_{\pm 1.46}$ & $63.72_{\pm 1.80}$ & $\boldsymbol{28.49_{\pm 17.61}}$ & $75.03_{\pm 4.14}$ & \underline{$77.12_{\pm 1.43}$} & \underline{$79.19_{\pm 2.45}$} & \underline{$63.66$} \\
            \textbf{PISA} (ours) & $\boldsymbol{79.88_{\pm 1.60}}$ & $\boldsymbol{67.12_{\pm 1.10}}$ & $\boldsymbol{66.20_{\pm 1.35}}$ & $43.80_{\pm 1.25}$ & $\boldsymbol{65.48_{\pm 1.70}}$ & \underline{$28.10_{\pm 11.50}$} & $\boldsymbol{77.10_{\pm 2.10}}$ & $\boldsymbol{79.45_{\pm 1.60}}$ & $\boldsymbol{81.50_{\pm 2.10}}$ & $\boldsymbol{65.40}$ \\
            \bottomrule
        \end{tabular}
    }
    \label{tab:results_realistic}
    \vspace{-0.6em}
\end{table}

Tables~\ref{tab:results_synthetic} and~\ref{tab:results_realistic} show that PISA attains the best average in both regimes: $83.37$ vs.\ $80.41$ on synthetic ($+2.96$) and $65.40$ vs.\ $63.66$ on realistic shifts ($+1.74$), against SuGAr as the strongest prior method. PISA is best on $6/6$ synthetic and $7/9$ realistic datasets; on the remaining two it is second (CMNIST) and third (SST5). The largest single gain, $+5.90$ points, is on SPMotif-$0.33$. On $5$ of the $9$ realistic datasets PISA's mean minus one standard deviation still exceeds the best baseline's mean; it falls marginally short on Ki-Assay and Twitter, and clearly short on SST5 and CMNIST.

\paragraph{Compute-matched control} To isolate gains from increased model count, we uniformly ensemble ten independently trained CIGA models (CIGA$\times$10). This raises CIGA from $57.83$ to $68.96$ on synthetic shifts and from $60.17$ to $61.26$ on realistic shifts, but remains well below PISA ($83.37$ and $65.40$, respectively). Thus, increased ensemble capacity alone does not explain PISA's gains.

\paragraph{Failure cases} On Graph-SST5, ERM, CIGA, SuGAr, and PISA all perform similarly ($44.21$, $44.20$, $43.38$, and $43.80$), suggesting an upstream extraction limitation rather than an aggregation-specific failure. Its degree-based shift acts on global graph structure, while sentiment information is distributed across semantic features and linguistic dependencies. CMNIST instead exhibits a strong attribute-level color shift with label noise; because color is distributed across node attributes rather than localized in a removable subgraph, it is also less aligned with the invariant-subgraph assumption. Consistently, CIGA reaches only $10.44$, while SuGAr and PISA obtain $28.49$ and $28.10$. We therefore view both as regimes where improved upstream extraction is needed.

\subsection{Does the Gain Survive Multiple Causal Subgraphs? (RQ2)}

SUMotif requires two causal motifs per label, while SPMotif provides a single-motif control. Averaged across biases, CIGA, SuGAr, and PISA drop by $5.22$, $5.18$, and $6.50$ points from SPMotif to SUMotif, respectively, while PISA's margin over SuGAr remains positive in both regimes ($+3.62$ and $+2.30$). This supports a narrow interpretation: PISA improves aggregation rather than extraction and therefore does not eliminate the harder multi-subgraph extraction problem.

The pattern is stronger on DrugOOD: relative to ERM, mean ROC-AUC improves by $+0.86$ for CIGA, $+2.76$ for SuGAr, and $+5.07$ for PISA, showing successive gains from multiple-subgraph extraction and prioritized aggregation.

\subsection{Number of Branches (RQ3)}\label{sec:ablation_branches}

We vary $n\in\{2,5,10,15\}$ on SPMotif-0.6 (Fig.~\ref{fig:ablation_branches}). PISA improves from $67.50$ at $n=2$ to a peak of $87.12$ at $n=10$, then falls to $78.80$ at $n=15$, motivating our default $n=10$. PISA also outperforms SuGAr at every tested $n$, indicating that its aggregation advantage is not specific to the default branch count. Computational scaling with $n$, including parameter counts, training times, and inference times, is reported in Appendix~\ref{app:compute_scaling}.

\subsection{Aggregator Capacity (RQ4)}\label{sec:ablation_mlp_layers}

\begin{figure}[t]
    \centering

    % Left: MLP depth
    \begin{subfigure}[c]{0.32\linewidth}
        \begin{minipage}[b][3.0cm][c]{\linewidth}
            \centering
            \includegraphics[width=\linewidth]{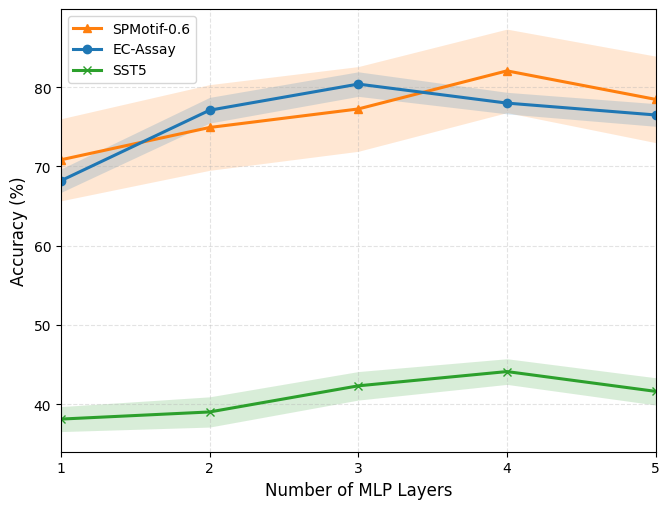}
        \end{minipage}
        \caption{MLP depth.}
        \label{fig:ablation_mlp_layers}
    \end{subfigure}
    \hfill
    % Middle: number of branches
    \begin{subfigure}[c]{0.32\linewidth}
        \begin{minipage}[b][3.0cm][c]{\linewidth}
            \centering
            \includegraphics[width=\linewidth]{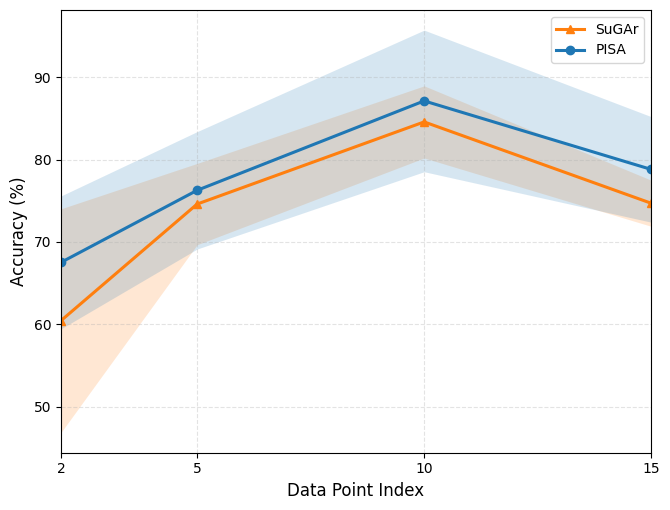}
        \end{minipage}
        \caption{Number of branches $n$.}
        \label{fig:ablation_branches}
    \end{subfigure}
    % \begin{subfigure}[c]{0.4\linewidth}
    %     \centering
    %     \begin{minipage}[b][3.0cm][c]{\linewidth}
    %         \resizebox{\linewidth}{!}{
    %             \begin{tabular}{lcccc}
    %                 \toprule
    %                 Method & $n=2$ & $n=5$ & $n=10$ & $n=15$ \\
    %                 \midrule
    %                 SuGAr
    %                 & $60.42_{\pm 13.59}$
    %                 & $74.59_{\pm 4.94}$
    %                 & $84.57_{\pm 4.38}$
    %                 & $74.69_{\pm 2.80}$ \\
    %                 PISA
    %                 & $\mathbf{67.50}_{\pm 8.07}$
    %                 & $\mathbf{76.26}_{\pm 7.13}$
    %                 & $\mathbf{87.12}_{\pm 8.60}$
    %                 & $\mathbf{78.80}_{\pm 6.41}$ \\
    %                 \bottomrule
    %             \end{tabular}
    %         }
    %     \end{minipage}
    %     \subcaption{Number of branches $n$.}
    %     \label{tab:ablation_branches}
    % \end{subfigure}
    \hfill
    % Right: parameter sharing
    \begin{subfigure}[c]{0.32\linewidth}
        \begin{minipage}[b][3.0cm][c]{\linewidth}
            \centering
            \includegraphics[width=\linewidth]{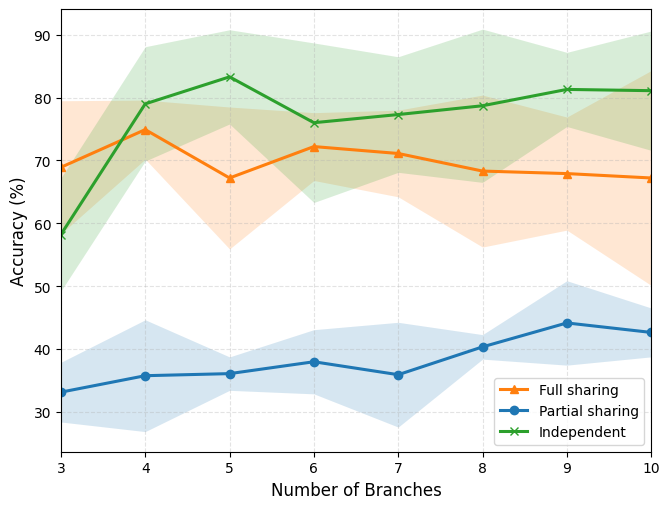}
        \end{minipage}
        \caption{Parameter sharing.}
        \label{fig:ablation_parameter_sharing}
    \end{subfigure}

    \caption{
        Ablation studies on SPMotif-0.6.
        Left: aggregator depth.
        Middle: effect of the number of branches $n$.
        Right: parameter sharing.
    }
    \label{fig:ablation_studies}
    \vspace{-0.8em}
\end{figure}

We sweep the depth of $A_\theta$ from $1$ to $5$ layers with everything else fixed (Fig.~\ref{fig:ablation_mlp_layers}). Accuracy rises then falls, peaking at $3$--$4$ layers---exactly the shape Remark~\ref{rem:depth} anticipates. A single linear layer realizes only a fixed weighting and cannot express instance-conditional prioritization at all; three to four layers suffice for soft gating and the suppression of redundant branches over a $30$-dimensional input; beyond that, capacity buys nothing the task requires while $\mathfrak{C}(\mathcal{A})$ in Eq.~\eqref{eq:decomp} keeps growing.

Prop.~\ref{prop:robust} rules out increased environment sensitivity as the cause, since its bound is uniform in $\mathcal{A}$. We instead hypothesize that the degradation reflects estimation-side effects, such as overfitting branch-logit noise or poorer calibration \citep{guo2017calibration}; distinguishing these mechanisms is left to future work.

\subsection{Parameter Sharing (RQ5)}

Full sharing helps only for small $n$ and falls below independent branches thereafter, consistent with shared representations limiting branch diversity. Partial sharing also underperforms independent branches at every tested $n$, suggesting that branch-specific attention over a shared backbone does not sufficiently decouple the predictors. More flexible approaches such as low-rank branch adapters remain promising directions for reducing Phase-I cost.

\section{Related Work}

\paragraph{Invariant and graph OOD learning} IRM \citep{irm}, IB-IRM \citep{ibirm}, V-REx \citep{rex}, EIIL \citep{eiil}, and GroupDRO \citep{groupdro} seek environment-stable predictors but do not specifically model graph structural shift. Graph methods instead learn invariant substructures, including GSAT \citep{gsat}, DIR \citep{dir}, GIB \citep{gib}, and CIGA \citep{ciga}. SuGAr \citep{sugar} is closest to PISA: it extracts several diverse subgraphs but combines them uniformly or greedily; our contribution targets this aggregation step.

\paragraph{Ensembles} Phase II relates to stacked generalization \citep{wolpert1992stacked,breiman1996stacked}, mixtures of experts \citep{jacobs1991adaptive,shazeer2017outrageously}, and deep ensembles \citep{lakshminarayanan2017simple}, while Prop.~\ref{prop:minvar} follows classical forecast combination \citep{bates1969combination}. Unlike parameter-space averaging \citep{swa,swad,diwa}, PISA combines frozen invariant predictors and establishes conditions under which the learned combination preserves their invariance.

\section{Limitations}
\label{sec:limitations}

\paragraph{Compute} PISA inherits Phase I from SuGAr and trains $n$ GNN branches; with $n=10$, this is substantially more expensive than a single CIGA model. To control for this additional Phase-I compute, we evaluate CIGA$\times$10, an ensemble of ten independently trained CIGA models with uniform prediction averaging. CIGA$\times$10 remains well below PISA on both synthetic (68.96 vs. 83.37) and realistic shifts (61.26 vs. 65.40), indicating that increased model count alone does not explain PISA's gains.

\paragraph{Scope of the guarantees} Prop.~\ref{prop:closure} assumes branches are \emph{exactly} invariant, which no trained featurizer is. Prop.~\ref{prop:robust} is conditional on the residual environment dependence $\delta$ of the learned branch outputs. PISA neither estimates nor optimizes this quantity; the proposition only shows that, once Phase I produces branch outputs with residual dependence $\delta$, arbitrary measurable aggregation does not amplify the corresponding worst-case environment gap. Assumption~\ref{assump:exog} further excludes shift in the marginal law of $C$, so covariate shift on the causal substructure is outside the analysis.

\paragraph{Extraction regime} PISA inherits the limitations of invariant-subgraph extraction: it is less effective when predictive information is distributed across attributes or global structure rather than localized in stable causal subgraphs, as illustrated by CMNIST and Graph-SST5.

\paragraph{Fixed branch count} The flattened aggregator ties $A_\theta$ to a specific $n$ and branch ordering, so branches cannot be added or dropped at test time without refitting Phase~II. A permutation-invariant encoder \citep{zaheer2017deep} would lift this at the cost of the branch-identity information Sec.~\ref{sec:method} argues is useful; we did not evaluate that trade-off.

\section{Conclusion}

Invariant learning on graphs has concentrated on \emph{finding} causal subgraphs and treated \emph{combining} them as an afterthought. This is a mistake with precise structure: uniform averaging is optimal only under a knife-edge balance of branch error covariances---exactly what the diversity regularizers behind multi-subgraph extraction exist to destroy---yet a learned combiner is nonetheless safe, because frozen invariant branches expose no environment-dependent information. PISA acts on this, replacing the combination rule with a small MLP over cached branch logits and improving synthetic and realistic OOD averages by $2.96$ and $1.74$ points at negligible cost; the residual multi-subgraph gap points to extraction as the next bottleneck.

\section*{Acknowledgments}
Maria Sofia Bucarelli acknowledges the French government National Research Agency (ANR) through the UCA JEDI (ANR-15-IDEX-0001),  through the EUR DS4H (ANR-17-EURE-0004), and through the 3IA Cote d'Azur Investments in the project with the reference number ANR-23-IACL-0001. Fabrizio Silvestri and acknowledge the project SAP\_RICERCA\_2025\_EVOLVE\_SILVES\_F\_01.

% For natbib users:
\bibliographystyle{unsrtnat}
\bibliography{reference}
% For bibLaTeX users:
% \printbibliography

\appendix
\section{Proofs}
\label{sec:proofs}

\subsection{Proof of Proposition~\ref{prop:minvar}}

All expectations are over the branch training randomness $\theta$ with $G$ fixed. Since $\one^\top w=1$ and $\mathbb{E}[\varepsilon^{(i)}\mid G]=0$, the combination is conditionally unbiased,
$\mathbb{E}[\widehat\eta_w\mid G]=\eta(G)$, and
\[
\mathbb{E}\big[(\widehat\eta_w-\eta)^2\mid G\big]
=\mathbb{E}\Big[\big(\textstyle\sum_i w_i\varepsilon^{(i)}\big)^2\,\Big|\,G\Big]
=w^\top\Sigma(G)\,w .
\]
Minimizing the quadratic form subject to $\one^\top w=1$ with a Lagrange multiplier $\lambda$ gives the stationarity condition $2\Sigma w=\lambda\one$, hence $w\propto\Sigma^{-1}\one$; normalizing to satisfy the constraint yields $w^\star$ in Eq.~\eqref{eq:wstar}, and it is the global minimum because $\Sigma\succ0$ makes the objective strictly convex on the affine constraint set. Substituting back,
\[
(w^\star)^\top\Sigma w^\star
=\frac{\one^\top\Sigma^{-1}\Sigma\,\Sigma^{-1}\one}{(\one^\top\Sigma^{-1}\one)^2}
=\frac{1}{\one^\top\Sigma^{-1}\one}.
\]
For the uniform rule, $\bar w^\top\Sigma\bar w=\one^\top\Sigma\one/n^2$, and Eq.~\eqref{eq:excess} follows with $\Delta(G)\ge0$ by optimality of $w^\star$. Equality holds iff $\bar w$ is itself a minimizer, i.e.\ iff $\one/n\propto\Sigma^{-1}\one$, equivalently $\Sigma(G)\one=\lambda\one$ for some $\lambda>0$; positivity of $\lambda$ follows from $\lambda=\one^\top\Sigma\one/n>0$. \hfill$\square$

\medskip
\noindent\emph{Proof of Corollary~\ref{cor:diversity}.} Write $(\Sigma\one)_i=\sum_k\Sigma_{ik}$. If $\Sigma_{ik}=\Sigma_{jk}$ for all $k\notin\{i,j\}$ and $\Sigma_{ii}\ne\Sigma_{jj}$, then
$(\Sigma\one)_i-(\Sigma\one)_j=\Sigma_{ii}-\Sigma_{jj}\ne0$ (the shared cross-term $\Sigma_{ij}=\Sigma_{ji}$ appears in both sums and cancels), so $\Sigma\one$ is not a multiple of $\one$ and Prop.~\ref{prop:minvar} gives $\Delta(G)>0$. \hfill$\square$

\medskip
\noindent\emph{The equal-covariance hypothesis cannot be dropped.} Heteroscedasticity by itself does not imply $\Delta(G)>0$, because unequal variances may be compensated by unequal covariances. Concretely, let $v_1=\one/\sqrt3$, $v_2=(1,-1,0)^\top/\sqrt2$, $v_3=(1,1,-2)^\top/\sqrt6$ and
\[
\Sigma=3\,v_1v_1^\top+1\,v_2v_2^\top+4\,v_3v_3^\top
=\begin{pmatrix} 13/6 & 7/6 & -1/3\\ 7/6 & 13/6 & -1/3\\ -1/3 & -1/3 & 11/3\end{pmatrix}.
\]
Its spectrum is $\{1,3,4\}$, so $\Sigma\succ0$; its diagonal $(13/6,13/6,11/3)$ is not constant, so the branches are heteroscedastic; yet $\one$ is an eigenvector, $\Sigma\one=3\,\one$, so by Prop.~\ref{prop:minvar} $\Delta=0$ and uniform averaging is exactly optimal. This is why Cor.~\ref{cor:diversity} is stated with the equal-covariance hypothesis rather than as a claim about heteroscedasticity alone.

\subsection{Proof of Proposition~\ref{prop:suff}}

For any measurable $A$ with $\mathbb{E}\|A(\Pmat)\|^2<\infty$, the orthogonality of the conditional expectation gives
\[
\mathbb{E}\big[\|A(\Pmat)-Y\|^2\big]
=\mathbb{E}\big[\|A(\Pmat)-\mathbb{E}[Y\mid\Pmat]\|^2\big]+\mathbb{E}\big[\|Y-\mathbb{E}[Y\mid\Pmat]\|^2\big],
\]
because the cross-term vanishes after conditioning on $\Pmat$. The second summand does not depend on $A$ and the first is minimized, at value $0$, by $A=A^\star$. Every static rule $A_w(\Pmat)=w^\top\Pmat$ is measurable, hence lies in $\mathcal{A}_{\mathrm{all}}$, and the domination claim follows. \hfill$\square$

\subsection{Proof of Proposition~\ref{prop:closure}}

By Assumption~\ref{assump:scm_graph_generation}, $\Gc=f^{\Gc}_{\mathrm{gen}}\big(f^{(A,X)^c}_{\mathrm{gen}}(C)\big)=:\kappa(C)$ for a measurable $\kappa$. By Def.~\ref{def:invariant} each branch satisfies $g^i(G)=\gamma^i(\Gc)$, so
\[
p^{(i)}=f^i_c\big(\gamma^i(\kappa(C))\big)=:\pi^i(C),
\]
a measurable function of $C$ alone. Hence $\Pmat=\big[\pi^1(C);\dots;\pi^n(C)\big]$ and, for any measurable $A$,
$\widehat{Y}=A(\Pmat)=:\Phi_1(C)$. By Assumption~\ref{assump:exog}, $Y=f_{\mathrm{inv}}(C)=:\Phi_2(C)$. Therefore
$(\widehat{Y},Y)=\big(\Phi_1(C),\Phi_2(C)\big)$ is a measurable function of $C$, and since $C\indep E$, measurability preserves independence: $(\widehat{Y},Y)\indep E$.

Consequently the conditional law $\mathbb{P}^e(Y\mid\widehat{Y})$ is the same for every $e$, and
\[
R^e(A)=\mathbb{E}\big[\ell(\widehat{Y},Y)\mid E=e\big]
=\mathbb{E}\big[\ell(\Phi_1(C),\Phi_2(C))\big],
\]
which contains no dependence on $e$. \hfill$\square$

\subsection{Proof of Proposition~\ref{prop:robust}}

Fix a measurable $A$ and let $\ell_A(\Pmat,Y)\coloneqq\ell(A(\Pmat),Y)$, so $0\le\ell_A\le B$. For any two probability measures $\mu,\nu$ on the same space and any measurable $f$ with $\|f\|_\infty\le B$,
\[
\Big|\textstyle\int f\,d\mu-\int f\,d\nu\Big|\;\le\;2B\,d_{\mathrm{TV}}(\mu,\nu),
\]
by the variational characterization $d_{\mathrm{TV}}(\mu,\nu)=\tfrac12\sup_{\|f\|_\infty\le1}|\int f d\mu-\int f d\nu|$. Applying this with $f=\ell_A$, $\mu=\mathbb{P}^e_{(\Pmat,Y)}$ and $\nu=\mathbb{P}^{e'}_{(\Pmat,Y)}$ gives
\[
\big|R^e(A)-R^{e'}(A)\big|\le 2B\,d_{\mathrm{TV}}\big(\mathbb{P}^e_{(\Pmat,Y)},\mathbb{P}^{e'}_{(\Pmat,Y)}\big)\le 2B\delta .
\]
The bound holds for each $A$ with the same constant $\delta$, which depends only on the branch outputs; taking the supremum over $e,e'$ gives Eq.~\eqref{eq:gap}. If every branch is exactly invariant then, by the proof of Prop.~\ref{prop:closure}, $(\Pmat,Y)$ is a function of $C$ and $C\indep E$, so the joint law of $(\Pmat,Y)$ is identical across environments and $\delta=0$. \hfill$\square$

\section{Full Structural Causal Models on Graph Generation}
\label{sec:graph_gen_process}

\begin{figure}[ht]
    \centering
    \begin{subfigure}{0.308\linewidth}
        \centering
        \includegraphics[width=\linewidth]{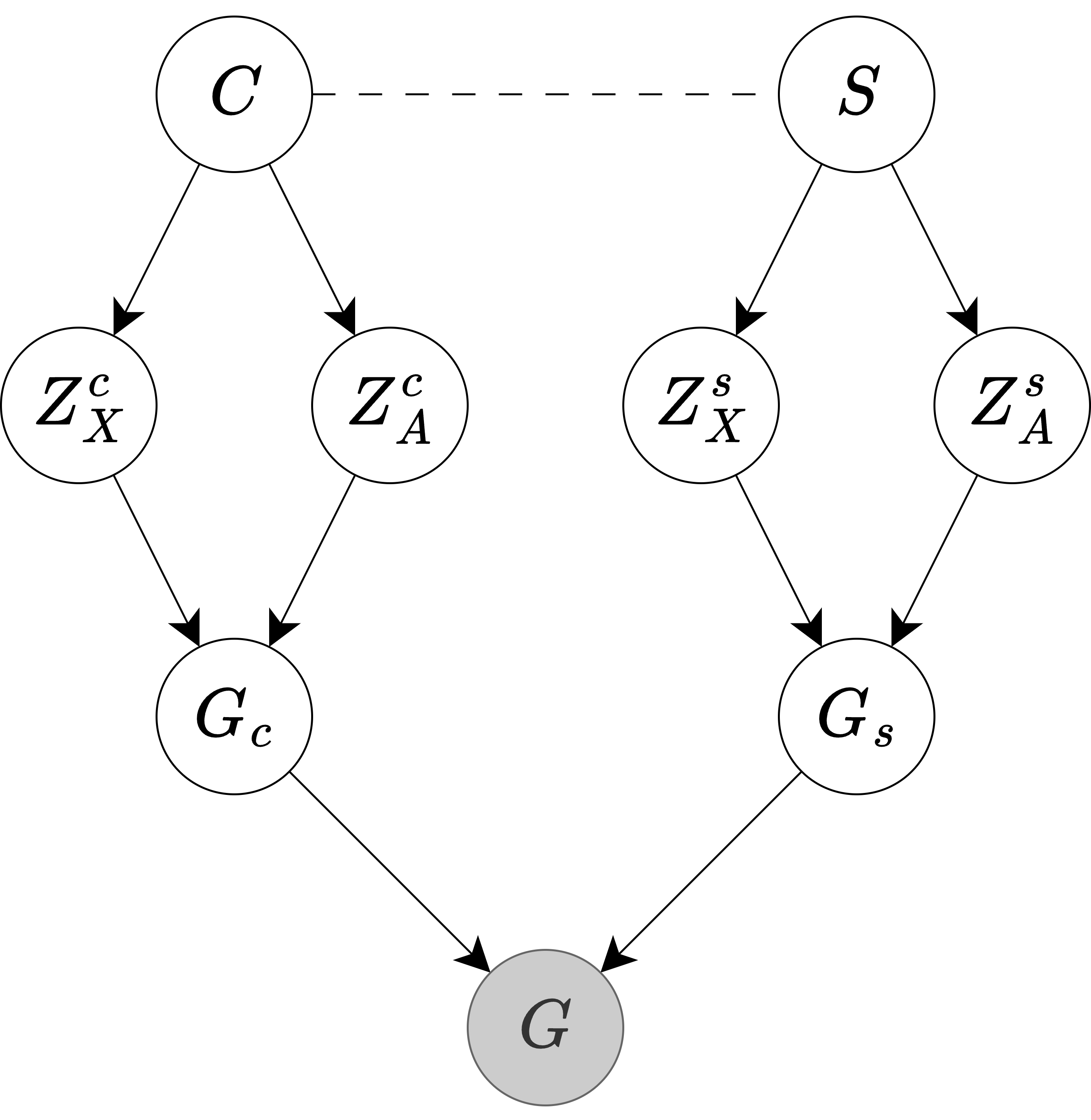}
        \caption{Graph generation SCM}
        \label{fig:scm_graph_generation}
    \end{subfigure}
    \hfill
    \begin{subfigure}{0.22\linewidth}
        \centering
        \includegraphics[width=\linewidth]{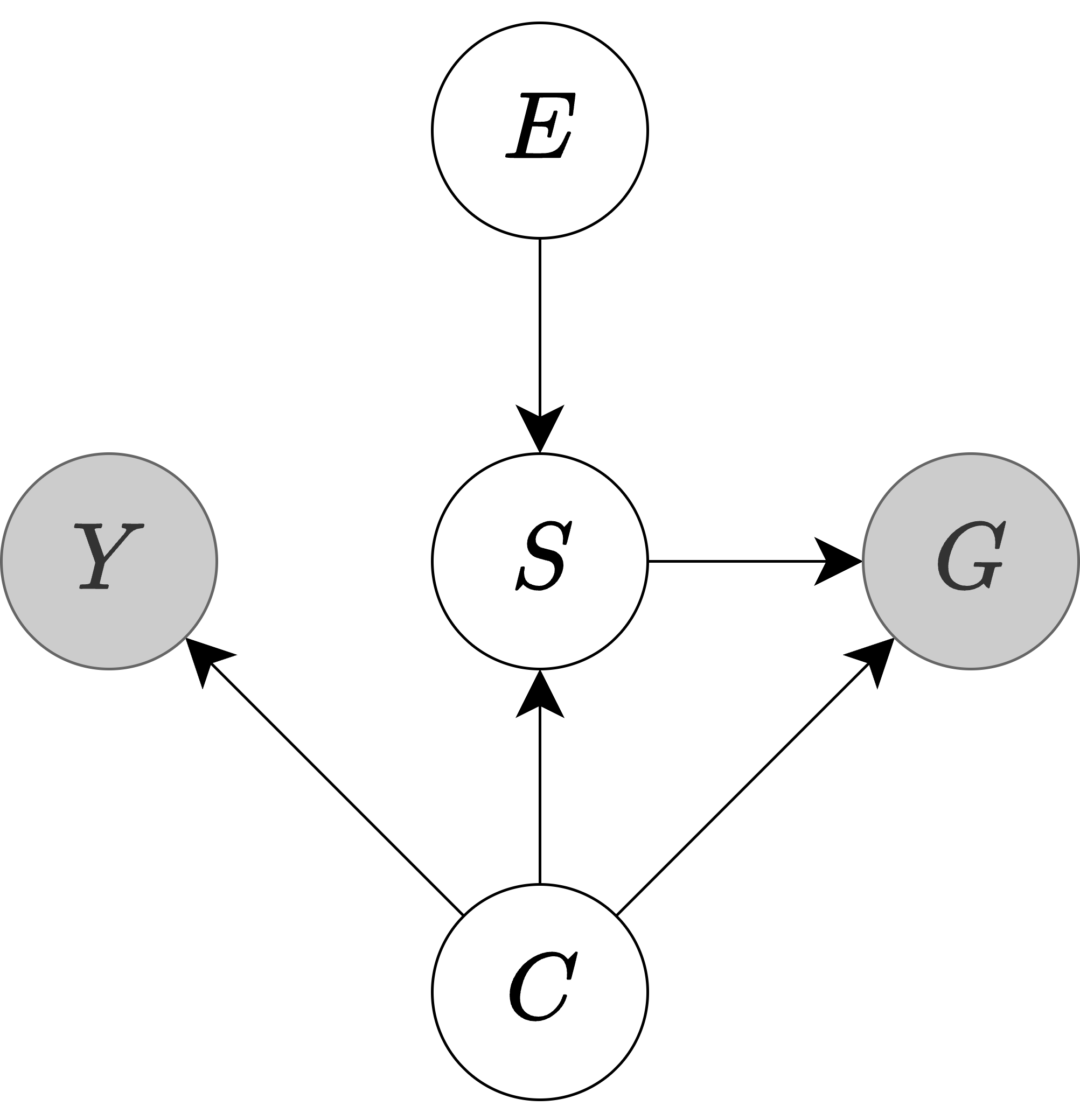}
        \caption{FIIF SCM}
        \label{fig:scm_fiif}
    \end{subfigure}
    \hfill
    \begin{subfigure}{0.22\linewidth}
        \centering
        \includegraphics[width=\linewidth]{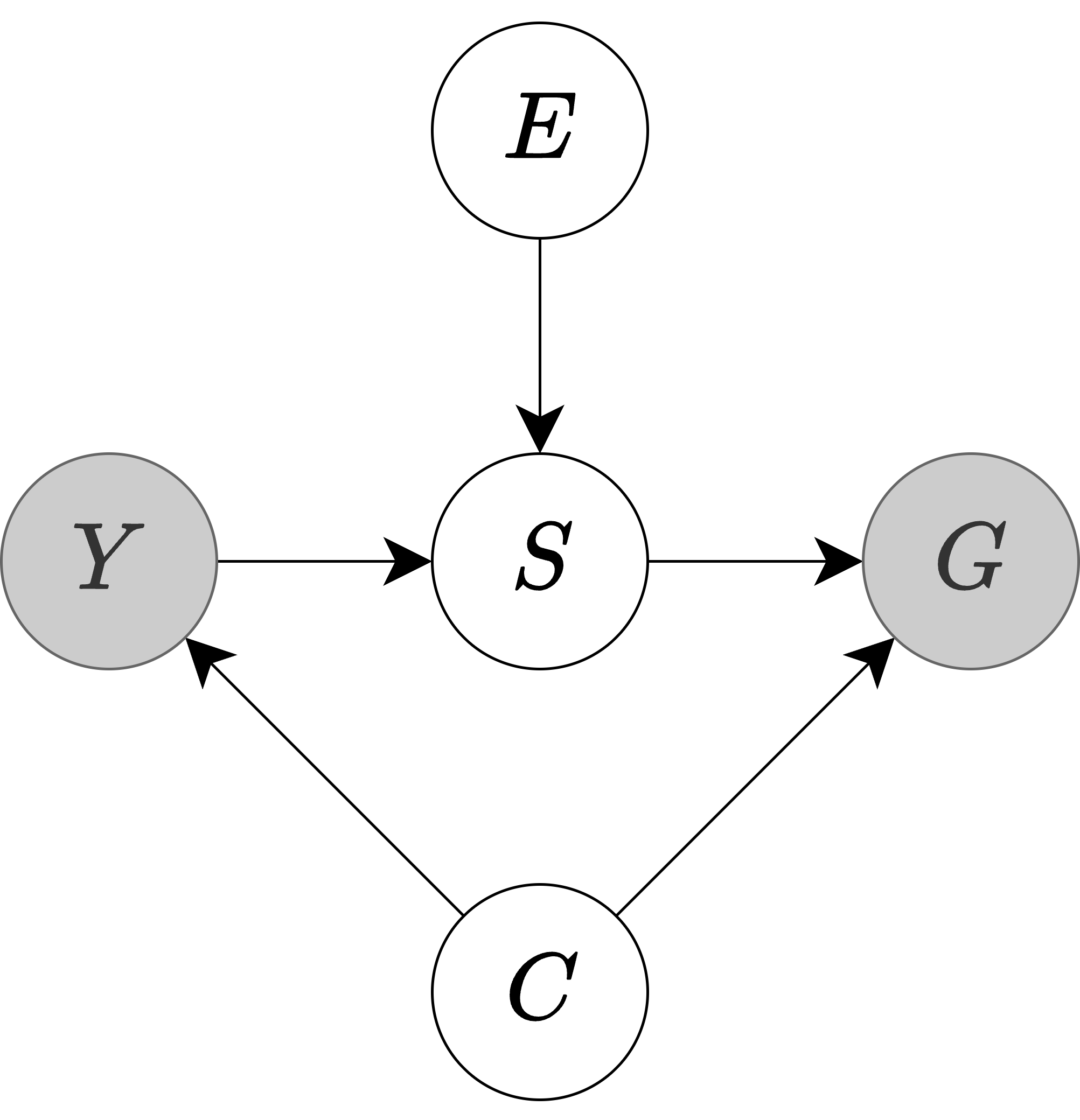}
        \caption{PIIF SCM}
        \label{fig:scm_piif}
    \end{subfigure}\\[2ex]
    \begin{subfigure}{0.308\linewidth}
        \centering
        \includegraphics[width=\linewidth]{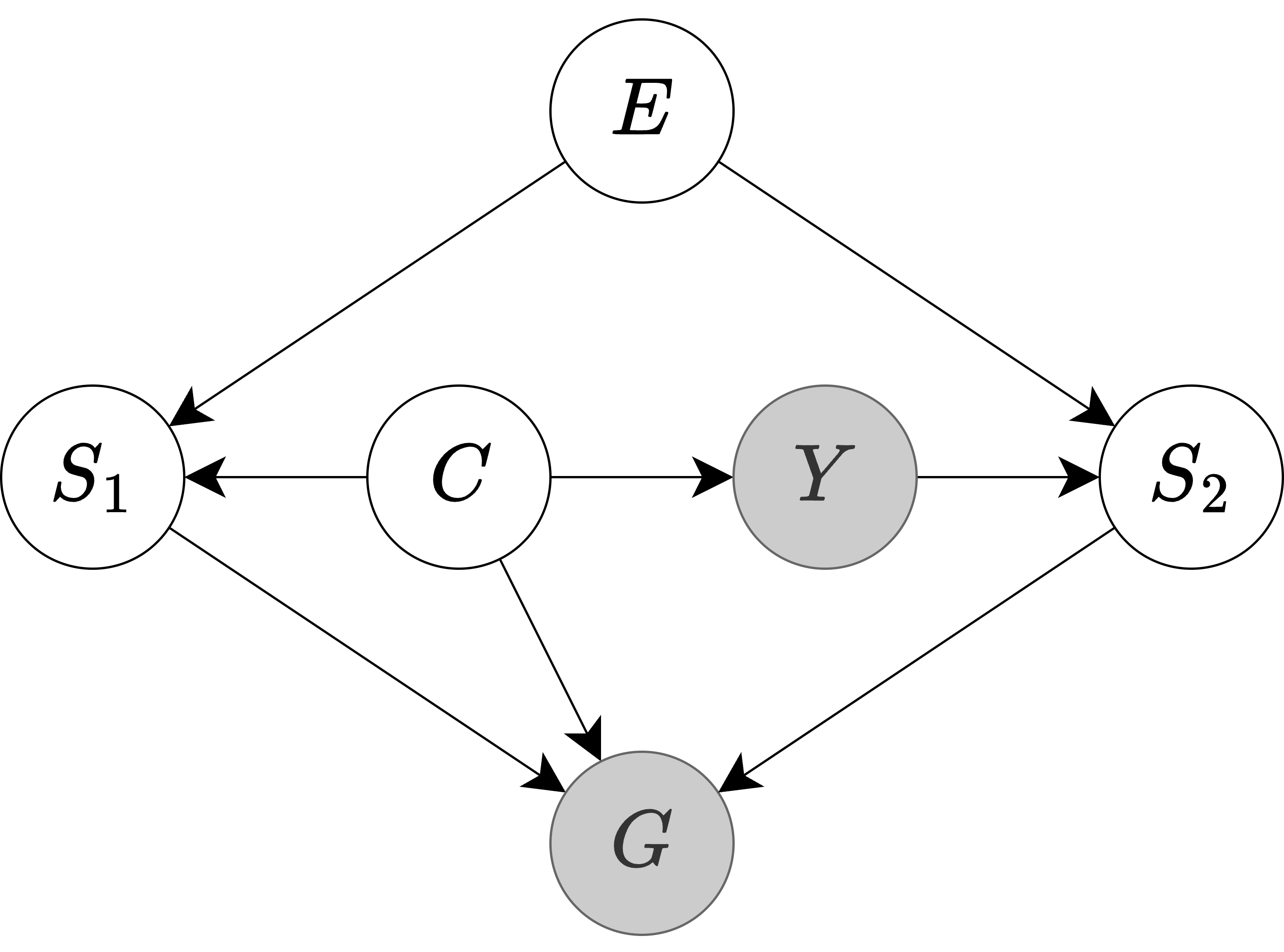}
        \caption{MIIF SCM}
        \label{fig:scm_miif}
    \end{subfigure}
    \caption{Full SCMs for graph distribution shifts. Each panel depicts a distinct interaction pattern between invariant ($C$) and spurious ($S$) latent factors and the environment $E$. In all panels $E$ has no directed path into $C$, which is the content of Assumption~\ref{assump:exog}.}
    \label{fig:scm}
\end{figure}

We present the full SCMs for the graph generation process summarized in Fig.~\ref{fig:scm}. These formalize how invariant and environment-dependent latent factors interact to produce observed graphs and labels under distribution shift. Precise formulations are given in Assumptions~\ref{assump:scm_graph_generation}, \ref{assump:scm_fiif}, \ref{assump:scm_piif} and \ref{assump:scm_miif}.

We adopt a latent-variable perspective and assume graphs are generated by a mapping $f_\text{gen}:\mathcal{Z}\to\mathcal{G}$, where the latent space is $\mathcal{Z}\subseteq\mathbb{R}^n$ and the graph space is
\[
\mathcal{G}=\bigcup_{N=1}^{\infty}\left(\{0,1\}^{N\times N}\times\mathbb{R}^{N\times d}\right),
\]
i.e.\ all graphs with binary adjacencies and $d$-dimensional node features. Let $E$ denote environments that may influence the data-generating process. Following prior work \citep{von2021self,ibirm}, we partition $Z\in\mathcal{Z}$ into an \emph{invariant} component $C\in\mathcal{C}=\mathbb{R}^{n_c}$ and an \emph{environment-varying} component $S\in\mathcal{S}=\mathbb{R}^{n_s}$, with $n=n_c+n_s$, according to whether the component is affected by $E$.

This decomposition mirrors the content/style separation used for images, with $E$ capturing factors such as capture location or device \citep{terraincognita,zhang2021causaladv,von2021self}. In graphs, $C$ controls invariant structural causes such as functional groups in molecules or domain-invariant motifs and communities, which are not altered by environments like scaffolds or experimental assays \citep{drugood}. In contrast, $S$ aggregates environment-specific variability (assay-dependent noise, context-specific feature biases). Crucially, $C$ and $S$ can interact with each other, with $E$, and with $Y$ in multiple ways at the latent level, inducing different types of spurious correlation \citep{ibirm}.

\begin{assumption}{Graph Generation SCM}\label{assump:scm_graph_generation}
    \begin{align*}
        &(Z^c_A, Z^c_X) \coloneqq f^{(A, X)^c}_\text{gen}(C), \qquad G_c \coloneqq f^{G_c}_\text{gen}(Z^c_A, Z^c_X),\\
        &(Z^s_A, Z^s_X) \coloneqq f^{(A, X)^s}_\text{gen}(S), \qquad G_s \coloneqq f^{G_s}_\text{gen}(Z^s_A, Z^s_X),\\
        &G \coloneqq f^G_\text{gen}(G_c, G_s).
    \end{align*}
\end{assumption}

The process is depicted in Fig.~\ref{fig:scm_graph_generation}. We decompose $f_\text{gen}$ into $f^{(A,X)^c}_\text{gen}$ and $f^{G_c}_\text{gen}$, generating the invariant subgraph $\Gc$ via latent adjacency- and feature-level variables $(Z^c_A,Z^c_X)$; and $f^{(A,X)^s}_\text{gen}$, $f^{G_s}_\text{gen}$, analogously generating $\Gs$ via $(Z^s_A,Z^s_X)$. The observed graph is the composition $f^G_\text{gen}(\Gc,\Gs)$, which may be a disjoint or overlapping join of $\Gc$ with one or several $\Gs$, or a more complex latent-controlled composition \citep{snijders1997estimation,lovasz2006limits,you2018graphrnn,luo2021graphdf,sizeinvariant}. Intuitively $Z^c_A,Z^s_A$ control structure-level properties (degrees, sizes, motif densities) while $Z^c_X,Z^s_X$ regulate attribute-level properties (homophily, feature distributions). Assumption~\ref{assump:scm_graph_generation} is deliberately compatible with many graph-generation families \citep{snijders1997estimation,lovasz2006limits,you2018graphrnn,luo2021graphdf}.

Because $E$ and $G$ are correlated through the generative mechanism, graphs from different environments exhibit distinct structure-level and feature-level properties \citep{knyazev2019understanding,local,sizeinvariant,chen2022understanding}. Depending on how the latents interact, these shifts become spuriously correlated with $Y$. We distinguish two axiomatic interaction types, FIIF and PIIF, and their mixture MIIF. Many OOD methods target only one type (GIB \citep{gib} and DIR \citep{dir} for FIIF; IRM \citep{irm} for PIIF), and failing to model either causes substantial degradation in practice \citep{aubin2021linear,nagarajan2020understanding}; our framework models both.

\begin{assumption}{FIIF SCM}\label{assump:scm_fiif}
    \[
    Y \coloneqq f_\text{inv}(C), \qquad S \coloneqq f_\text{spu}(C, E), \qquad G \coloneqq f_\text{gen}(C, S).
    \]
\end{assumption}

\begin{assumption}{PIIF SCM}\label{assump:scm_piif}
    \[
    Y \coloneqq f_\text{inv}(C), \qquad S \coloneqq f_\text{spu}(Y, E), \qquad G \coloneqq f_\text{gen}(C, S).
    \]
\end{assumption}

\begin{assumption}{MIIF SCM}\label{assump:scm_miif}
    \[
    Y \coloneqq f_\text{inv}(C), \qquad S_1 \coloneqq f_\text{spu}(C, E), \qquad S_2 \coloneqq f_\text{spu}(Y, E), \qquad G \coloneqq f_\text{gen}(C, S_1, S_2).
    \]
\end{assumption}

Interactions between $C$ and $S$ are categorized into \emph{Fully Informative Invariant Features} (FIIF; Fig.~\ref{fig:scm_fiif}) and \emph{Partially Informative Invariant Features} (PIIF; Fig.~\ref{fig:scm_piif}) depending on whether $C$ is sufficient for $Y$, i.e.\ $(S,E)\indep Y\mid C$; \emph{Mixed} interactions (MIIF; Fig.~\ref{fig:scm_miif}) blend both. We follow \citet{irm,ibirm} in the FIIF/PIIF formulation, omitting explicit noise terms for clarity \citep{causality,peters2017elements}. In every case $Y$ is a function of $C$ alone and $E$ enters only through $S$, which is what Assumption~\ref{assump:exog} records and what the proof of Prop.~\ref{prop:closure} uses. A standard separability condition formalizes that $C$ is more informative about $Y$ than $S$:

\begin{assumption}{Latent Separability}\label{assump:latent_sep}
    \[
    H(C \mid Y) \le H(S \mid Y),
    \]
\end{assumption}
where $H$ is the entropy. This captures the intuition that, conditional on the label, invariant content clusters more tightly than the spurious component \citep{burshtein1992minimum,chapelle2009semi,scholkopf2022causality,scholkopf2021toward}, a separation property exploited by classification objectives \citep{muller2018introduction,chen2005tutorial,mika1999fisher}.

\section{Algorithm}
\label{sec:algo}

\begin{algorithm}[H]
\caption{PISA}
\label{alg:pisa}
\begin{algorithmic}[1]
\Require training set $\mathcal{D}^{\mathrm{tr}}$; branches $n$; selection ratio $s_c$; drop rate $q$; weights $\alpha,\beta$
\Statex \textbf{Phase I --- diverse invariant extraction}
\State initialize $\{(g^i,f^i_c)\}_{i=1}^n$ identically
\For{each minibatch $\mathcal{B}\subset\mathcal{D}^{\mathrm{tr}}$}
  \For{$i=1,\dots,n$}
    \State $G^{(i)}\gets\textsc{DropEdges}(G,q)$;\quad $m^i\gets g^i(G^{(i)})$;\quad $\Ghc^i\gets\textsc{TopK}(m^i,s_c)$
    \State $p^{(i)}\gets f^i_c(\Ghc^i)$
  \EndFor
  \State update $\{(g^i,f^i_c)\}$ on $\mathcal{L}_{\mathrm{I}}$ of Eq.~\eqref{eq:phase1} using $\delta_{ij}$ from Eq.~\eqref{eq:div}
\EndFor
\Statex \textbf{Phase II --- prioritized aggregation}
\State freeze $\{(g^i,f^i_c)\}$; \textbf{precompute} the logit tensor $\{\Pmat(G_k)\}_{k=1}^{N}\in\mathbb{R}^{N\times n\times C}$ in one pass
\For{each epoch}\Comment{no GNN forward passes required}
  \State update $\theta$ on $\mathcal{L}_{\mathrm{II}}$ with $\widehat{Y}=A_\theta(\mathrm{vec}(\Pmat))$
\EndFor
\State \Return $\{(g^i,f^i_c)\}_{i=1}^n$, $A_\theta$
\end{algorithmic}
\end{algorithm}

\section{Notation}
\label{sec:notation}

\begin{center}
\begin{tabular}{ll}
\toprule
Symbol & Meaning \\
\midrule
$G,\;Y,\;E$ & observed graph, label, environment \\
$C,\;S$ & invariant (content) and environment-varying (style) latents \\
$\Gc,\;\Gs$ & invariant and spurious subgraphs generated from $C$ and $S$ \\
$\Ghc^i$ & candidate invariant subgraph proposed by branch $i$ \\
$g^i,\;f^i_c$ & featurizer and classifier of branch $i$ \\
$m^i\in[0,1]^{|E|}$ & soft edge-importance mask of branch $i$ \\
$n,\;C$ & number of branches, number of classes \\
$p^{(i)}\in\mathbb{R}^{C}$ & logits of branch $i$ \\
$\Pmat\in\mathbb{R}^{n\times C}$ & stacked branch logits, Eq.~\eqref{eq:stack} \\
$A_\theta$ & aggregator (MLP), Eq.~\eqref{eq:agg} \\
$\Sigma(G)$ & conditional covariance of branch residuals \\
$w^\star(G)$ & instance-optimal combination weights, Eq.~\eqref{eq:wstar} \\
$s_c,\;q$ & subgraph selection ratio, edge-drop probability \\
$\alpha,\;\beta$ & contrastive and diversity loss weights \\
\bottomrule
\end{tabular}
\end{center}

\section{Details of the Empirical Study}
\label{sec:app_emp}

\subsection{Datasets}

\subsubsection{Synthetic Datasets}

\paragraph{SPMotif.} A 3-class benchmark in which the task is to identify which of three motifs (\emph{House}, \emph{Cycle}, \emph{Crane}) is present. We generate \textbf{3{,}000} training graphs per class and \textbf{1{,}000} validation and \textbf{1{,}000} test graphs per class. Distribution shift is injected \emph{only} into the training split; validation and test are unbiased.

Each graph comprises (i) a \emph{motif} subgraph that deterministically sets the label and (ii) a \emph{base} graph inducing spurious correlation. To study structure-level shift we correlate each motif with one of three base types (\emph{Tree}, \emph{Ladder}, \emph{Wheel}): given bias $b$, a specific motif co-occurs with its designated base graph with probability $b$ and with each of the other two with probability $(1-b)/2$. Following \citet{ciga} we use $b\in\{0.33,0.6,0.9\}$. Node features are random, isolating structure-driven effects.

\paragraph{SUMotif.} Extends SPMotif to labels determined by \emph{multiple} subgraphs. Each of the three classes corresponds to a \emph{pair} of motifs from \{House, Cycle, Crane\}: \{House--Cycle, Cycle--Crane, Crane--House\}. Split sizes and the bias protocol match SPMotif. The two motif subgraphs are \emph{not} connected to each other; each attaches at random to the base graph. Node features are again random.

\subsubsection{Realistic Datasets}

We use six settings from the \textbf{DrugOOD} benchmark \citep{drugood} (Assay, Scaffold and Size splits for both EC50 and Ki), a graphified \textbf{ColoredMNIST} (\textbf{CMNIST-sp}) \citep{irm,knyazev2019understanding} for attribute shift, and two graph-of-text datasets, \textbf{Graph-SST5} and \textbf{Twitter}, with degree-based shift \citep{explainability}.

\paragraph{DrugOOD} \citet{drugood} provide an OOD benchmark for AI-aided drug discovery, predicting binding affinity between targets and ligands, curated from \textbf{ChEMBL} \citep{mendez2019chembl}. Shifts arise across assays, scaffolds and molecule sizes. We evaluate six ligand-based affinity tasks: \textit{lbap-core-ec50-\{assay,scaffold,size\}} and \textit{lbap-core-ki-\{assay,scaffold,size\}}. All data are used as released; see \citet{drugood} for preprocessing details.

\paragraph{CMNIST-sp.} The ColoredMNIST setup of \citet{irm}, converted to graphs via the superpixel pipeline of \citet{knyazev2019understanding}. Digits $0$--$4$ map to $y{=}0$ and $5$--$9$ to $y{=}1$, followed by label flips with probability $0.25$. During training (without environment stratification), images labeled $0$ (resp.\ $1$) receive green (resp.\ red) with average probability $0.15$; for validation and test the colour--label correlation is strengthened to $0.9$, producing an attribute-level shift.

\paragraph{Graph-SST datasets.} Following \citet{explainability} we convert sentence-level sentiment datasets \citep{socher2013recursive,dong2014adaptive} into graphs: node features from \textbf{BERT} \citep{devlin2019bert}, edges from a \textbf{Biaffine} parser \citep{gardner2018allennlp}. Degree-based shift is induced by splitting on average graph degree. For \textbf{Graph-SST5}, graphs below the 50th percentile form the training set, those between the 50th and 80th percentiles form validation, and the rest form test. For \textbf{Twitter} the split is \emph{reversed}, testing generalization from higher- to lower-degree graphs.

\subsection{Baselines}

From the Euclidean OOD literature: ERM \citep{principles}, IRM \citep{irm}, V-REx \citep{rex}, EIIL \citep{eiil}, IB-IRM \citep{ibirm}. From the graph domain: GREA \citep{grea}, GSAT \citep{gsat}, CAL \citep{cal}, GIL \citep{gil}, DisC \citep{disc}, CIGA \citep{ciga} and SuGAr \citep{sugar}.

For all methods relying on CIGA-style extraction we fix the selection ratio $s_c$ across base models for fairness. Each PISA base model is a CIGA instance trained on the full input graph, and PISA aggregates $n=10$ such models unless stated otherwise. Note that PISA and SuGAr share Phase~I exactly, so their comparison isolates the aggregation rule; comparisons against single-model baselines are not compute-matched, as discussed in Sec.~\ref{sec:limitations}.

CIGA$\times$10 is a compute-matched ensemble control consisting of ten independently trained CIGA models whose predictions are combined by uniform averaging. Unlike SuGAr/PISA, the branches are trained independently and therefore do not use explicit cross-branch diversity regularization. This isolates the benefit of simply increasing the number of invariant predictors from that of coordinated diverse extraction and learned aggregation.

\subsection{Implementation and Evaluation Details}

\paragraph{Global training configuration} PISA is trained in two stages. Phase I trains the invariant branches, after which the validation-selected branch checkpoints are frozen and Phase II trains only the aggregation network. Both phases use Adam; no learning-rate scheduler is used. Unless specified otherwise below, we use $n=10$ branches, a maximum of 40 epochs per phase, mean graph pooling, contrastive temperature $1.0$, $\mathcal{L}_2$-normalized contrastive representations, and an effective diversity-regularization coefficient of $10^{-2}$. The Phase-II aggregator uses hidden width 32.

\begin{table}[H]
    \caption{Dataset-specific hyperparameters used for PISA. Hidden dimensions are reported as branch encoder / branch classifier / aggregator.}
    \label{tab:pisa_hyperparameters}
    \centering
    \resizebox{\linewidth}{!}{
        \begin{tabular}{lcccccccc}
            \toprule
            Dataset & Backbone & LR & Batch & r & Contrast & Hidden dim. & Dropout & Agg. \\
            \midrule
            SPMotif & 3-layer GCN & $10^{-3}$ & 32 & .25 & 2 & 32/32/32 & 0 & 2-layer \\
            SUMotif & 3-layer GCN & $10^{-3}$ & 32 & .25 & 2 & 32/32/32 & 0 & 2-layer \\
            EC-Assay & 4-layer GIN & $10^{-3}$ & 128 & .80 & 8 & 32/32/32 & .1 & 3-layer \\
            EC-Scaffold & 4-layer GIN & $10^{-3}$ & 128 & .80 & 32 & 32/32/32 & .1 & 3-layer \\
            EC-Size & 4-layer GIN & $10^{-3}$ & 128 & .80 & 16 & 32/32/32 & .1 & 3-layer \\
            Ki-Assay & 4-layer GIN & $10^{-3}$ & 128 & .80 & 8 & 32/32/32 & .1 & 3-layer \\
            Ki-Scaffold & 4-layer GIN & $10^{-3}$ & 128 & .80 & 8 & 32/32/32 & .1 & 3-layer \\
            Ki-Size & 4-layer GIN & $10^{-3}$ & 128 & .80 & 8 & 32/32/32 & .1 & 3-layer \\
            Graph-SST5 & 3-layer GCN & $10^{-3}$ & 32 & .50 & 8 & 128/128/32 & .1 & 3-layer \\
            Twitter & 3-layer GCN & $10^{-3}$ & 32 & .60 & 8 & 128/128/32 & .1 & 3-layer \\
            CMNIST & 3-layer GCN & $10^{-3}$ & 32 & .80 & 32 & 32/128/32 & .1 & 2-layer \\
            \bottomrule
        \end{tabular}
    }
    \vspace{-0.4em}
\end{table}

Here, $r$ denotes the retained invariant-subgraph ratio; “Contrast” is the Phase-I contrastive-loss coefficient; hidden dimensions are reported as branch encoder / branch classifier / aggregator. All experiments use a sampling rate of $95\%$, diversity multiplier $1$, and $\mathrm{reg}=10^{-2}$.

\paragraph{Architectures} SPMotif and SUMotif use 3-layer GCN branch encoders; DrugOOD uses 4-layer GINs; Graph-SST5, Graph-Twitter and CMNIST use 3-layer GCNs. The two-layer aggregator used for SPMotif/SUMotif maps each class's ten branch logits as

\[\mathrm{Linear}(10,32)\rightarrow\mathrm{ReLU}\rightarrow\mathrm{Linear}(32,1),\]

and contains 385 trainable parameters. The three-layer variant adds one $\mathrm{Linear}(32,32)\rightarrow\mathrm{ReLU}$ hidden block and contains 1,441 trainable parameters.

\paragraph{Model selection} Each Phase-I branch is checkpointed independently at its highest validation score; Phase I does not use early stopping. The selected branch checkpoints are then frozen and used to train Phase II. The aggregator checkpoint is likewise selected by its highest validation score, with early stopping considered after 20 epochs and a patience of five non-improving epochs. The test set is never used for model selection; every reported test result corresponds to the validation-selected checkpoint.

\paragraph{Seeds and reporting} Each experiment is repeated over five independent initialization seeds $\{1,2,3,4,5\}$; for $n=10$, each run contains ten branch identifiers/seeds. We report the arithmetic mean and sample standard deviation ($N-1$) over the five independent runs. Accuracy is used for all datasets except DrugOOD, for which we report ROC-AUC. The AVG columns in Tables~\ref{tab:results_synthetic} and \ref{tab:results_realistic} are unweighted averages of the per-dataset means.

\subsection{Computational Scaling with the Number of Branches}\label{app:compute_scaling}

\begin{table}[H]
    \centering
    \caption{Computational scaling with the number of branches $n$ on
    SPMotif-0.6. Times are reported in seconds (mean $\pm$ std).}
    \label{tab:branch_cost}
    \small
    \setlength{\tabcolsep}{4pt}
    \begin{tabular}{lrrrr}
        \toprule
         & $n=2$ & $n=5$ & $n=10$ & $n=15$ \\
        \midrule
        Phase-I params.
        & 42,412 & 106,030 & 212,060 & 318,090 \\
        Phase-II params.
        & 129 & 225 & 385 & 545 \\
        Phase-I train (s)
        & $1083.21_{\pm 9.18}$
        & $2440.18_{\pm 33.78}$
        & $4571.75_{\pm 25.31}$
        & $6829.79_{\pm 53.71}$ \\
        Phase-II train (s)
        & $420.92_{\pm 146.51}$
        & $664.68_{\pm 69.87}$
        & $1117.40_{\pm 7.31}$
        & $1774.37_{\pm 398.55}$ \\
        Inference (s)
        & $1.78_{\pm 0.01}$
        & $3.21_{\pm 0.04}$
        & $5.60_{\pm 0.03}$
        & $7.93_{\pm 0.16}$ \\
        \bottomrule
    \end{tabular}
\end{table}

Phase-I parameter count, training time, and inference time increase approximately linearly with $n$, reflecting the cost of evaluating additional GNN branches. In contrast, the Phase-II aggregator remains lightweight: at the default $n=10$, it contains only 385 parameters, approximately $0.18\%$ of the 212,060 Phase-I parameters. Together with the branch-count ablation in Sec.~\ref{sec:ablation_branches}, where performance peaks at $n=10$ among the tested values, these results motivate $n=10$ as a practical performance--cost trade-off.

\section{Use of Generative AI}

Large language models were used for language editing, text correction and \LaTeX{} formatting, and as a drafting aid for the some formal statements and proofs. All of these were verified and corrected by the authors, who take full responsibility for the correctness of the paper's content.

\end{document}